%% file: bb_alpha_icml.tex
\documentclass{article}

\usepackage{times}
\usepackage{graphicx} % more modern
\usepackage{subfigure} 

\usepackage{url}
\usepackage{bm, amsmath}
\usepackage{wrapfig}

\usepackage{natbib}

\usepackage{algorithm}
\usepackage{algorithmic}

\newcommand{\ica}{\hspace{0.25cm}}

\usepackage{hyperref}

\usepackage[accepted]{icml2016} 

\icmltitlerunning{Black-Box Alpha}

\usepackage{wrapfig}
\newcommand{\bt}{\boldsymbol{\theta}}
\newcommand{\bx}{\mathbf{x}}
\begin{document} 

\twocolumn[
\icmltitle{Black-Box $\alpha$-Divergence Minimization}

\icmlauthor{Jos\'e Miguel Hern\'andez-Lobato$^{1*}$}{jmh@seas.harvard.edu}
\icmlauthor{Yingzhen Li$^{2*}$}{yl494@cam.ac.uk}
\icmlauthor{Mark Rowland$^2$}{mr504@cam.ac.uk}
\icmlauthor{Daniel Hern\'andez-Lobato$^3$}{daniel.hernandez@uam.es}
\icmlauthor{Thang D.~Bui$^2$}{tdb40@cam.ac.uk}
\icmlauthor{Richard E.~Turner$^2$}{ret26@cam.ac.uk}
\icmladdress{$^1$Harvard University, $^2$University of Cambridge, $^3$Universidad Aut\'onoma de Madrid, $^*$Both authors contributed equally.}
\icmlkeywords{boring formatting information, machine learning, ICML}

\vskip 0.3in
]

\begin{abstract} 
Black-box alpha (BB-$\alpha$) is a new approximate inference method based on
the minimization of $\alpha$-divergences.  BB-$\alpha$ scales to large datasets
because it can be implemented using stochastic gradient descent. BB-$\alpha$
can be applied to complex probabilistic models with little effort since it only
requires as input the likelihood function and its gradients. These gradients
can be easily obtained using automatic differentiation. By changing the
divergence parameter $\alpha$, the method is able to interpolate between
variational Bayes (VB) ($\alpha \rightarrow 0$) and an algorithm similar to expectation
propagation (EP) ($\alpha = 1$). Experiments on probit regression and neural
network regression and classification problems show that BB-$\alpha$ with
non-standard settings of $\alpha$, such as $\alpha = 0.5$, usually produces better
predictions than with $\alpha \rightarrow 0$ (VB) or $\alpha = 1$ (EP).
\end{abstract} 

% introduction
\input{intro}

% background
\input{section2}

% main idea
\input{section3}

% experiments
\input{experiments}

% conclusion
\input{conclusion}

\subsection*{Acknowledgements}

JMHL acknowledges support from the Rafael del Pino Foundation. YL
thanks the Schlumberger Foundation Faculty for the Future fellowship on
supporting her PhD study. MR acknowledges support from UK Engineering
and Physical Sciences Research Council (EPSRC) grant EP/L016516/1 for
the University of Cambridge Centre for Doctoral Training, the Cambridge
Centre for Analysis. TDB thanks Google for funding his European
Doctoral Fellowship. DHL acknowledge support from Plan National
I+D+i, Grant TIN2013-42351-P and TIN2015-70308-REDT, and from Comunidad
de Madrid, Grant S2013/ICE-2845 CASI-CAM-CM. RET thanks EPSRC grant
\#EP/L000776/1 and \#EP/M026957/1.

\nocite{langley00}

\bibliography{references.bib}
\bibliographystyle{icml2016}

\onecolumn
\input{appendix}

\end{document}

%% file: intro.tex
\section{Introduction}
Bayesian probabilistic modelling provides useful tools for making predictions
from data.  The formalism requires a probabilistic model
$p(\bm{x}|\bm{\theta})$, parameterized by a parameter vector $\bm{\theta}$,
over the observations $\mathcal{D} = \{\bm{x}_n\}_{n=1}^N$. Bayesian inference
treats $\bm{\theta}$ as a random variable and predictions are then made by averaging with respect to the posterior belief
%, obtained
%using  with Bayes' rule given the prior and the observations: 
\begin{equation*} 
p(\bm{\theta}|\mathcal{D}) \propto \left[ \prod_{n=1}^N p(\bm{x}_n|\bm{\theta}) \right] p_0(\bm{\theta})\,,
\end{equation*}
where $p(\bm{x}_n|\bm{\theta})$ is a likelihood factor and $p_0(\bm{\theta})$ is the prior.
Unfortunately the computation of this posterior distribution is often intractable for
many useful probabilistic models. 
One can use approximate inference techniques to sidestep this difficulty. Two
examples are variational Bayes (VB) \cite{jakkola99} and expectation
propagation (EP) \cite{minka2001}.  These methods adjust the parameters of a
tractable distribution so that it is close to the true posterior, by finding
an stationary point of an energy function. Both VB and EP are specific cases
of local $\alpha$-divergence minimization, where
the parameter 
$\alpha\in(-\infty,+\infty)\setminus\{0\}$ specifies the
properties of the divergence to be minimized \cite{minka2005}. If $\alpha \rightarrow 0$, VB is
obtained and $\alpha=1$ gives EP \cite{minka2005}. Power EP (PEP)
\cite{minka2004} extends EP to general settings of $\alpha$, whose optimal
value may be model, dataset and/or task specific.

%%%%%%%%% optimisation method: gradient descent or message passing %%%%%%%%
%%%%%%%%% problem of VI and when we want to use EP %%%%%%%%%%%%%%%%%%
EP can provide better solutions than VB in specific
cases. For instance, VB provides poor approximations when non-smooth
likelihood functions are used \cite{cunningham2011, turner+sahani2011, opper2005expectation}. 
EP also often performs better when factored approximations are employed \cite{Turner2011,minka2001thesis}.
%However", compared to VB, EP has received less attention because of the
%complexity of its implementation.  
%EP constructs a posterior approximation containing local
%approximations to factors in the likelihood function.
%Moreover, EP utilizes cheap
%local computations and message passing to obtain fixed points of the energy function, 
%while VI often directly optimizes the typically non-convex variational free energy with gradient descent.
%
%%%%%%%%% problem for local optimisation  %%%%%%%%%%
There are, however, issues that hinder the wide deployment of EP,
and by extension of power EP too.
First, EP
requires to store in memory local approximations of the likelihood factors. This has
prohibitive memory cost for large datasets and big models. Second, efficient
implementations of EP based on message passing have no convergence guarantees
to an stationary point of the energy function \cite{minka2001}. 

Previous work has addressed the first issue by using ``factor tying"/``local
parameter sharing" through the ``stochastic EP"/``averaged EP" (SEP/AEP) algorithms
\cite{li2015stochastic, dehaene2015}. 
%These works provided theoretical and
%empirical results showing that the approximations work almost as well as
%regular EP whilst offering substantial memory efficiencies. 
However, the energy
function of SEP/AEP is unknown, making the attempts for proving
convergence of these algorithms difficult. 
On the second issue,
\citet{heskes2002expectation} and \citet{opper2005expectation} derived a
convergent double-loop implementation of EP.
%algorithm based on the original min-max optimization
%problem. It can, 
However, it can be far slower than the original message passing procedure.
\citet{teh2015} proposed the stochastic natural-gradient EP (SNEP) method,
which is also double-loop-like, but in practice, they only perform one-step
inner loop update to speed up training.

\begin{figure}
\centering
\includegraphics[width=1.0\linewidth]{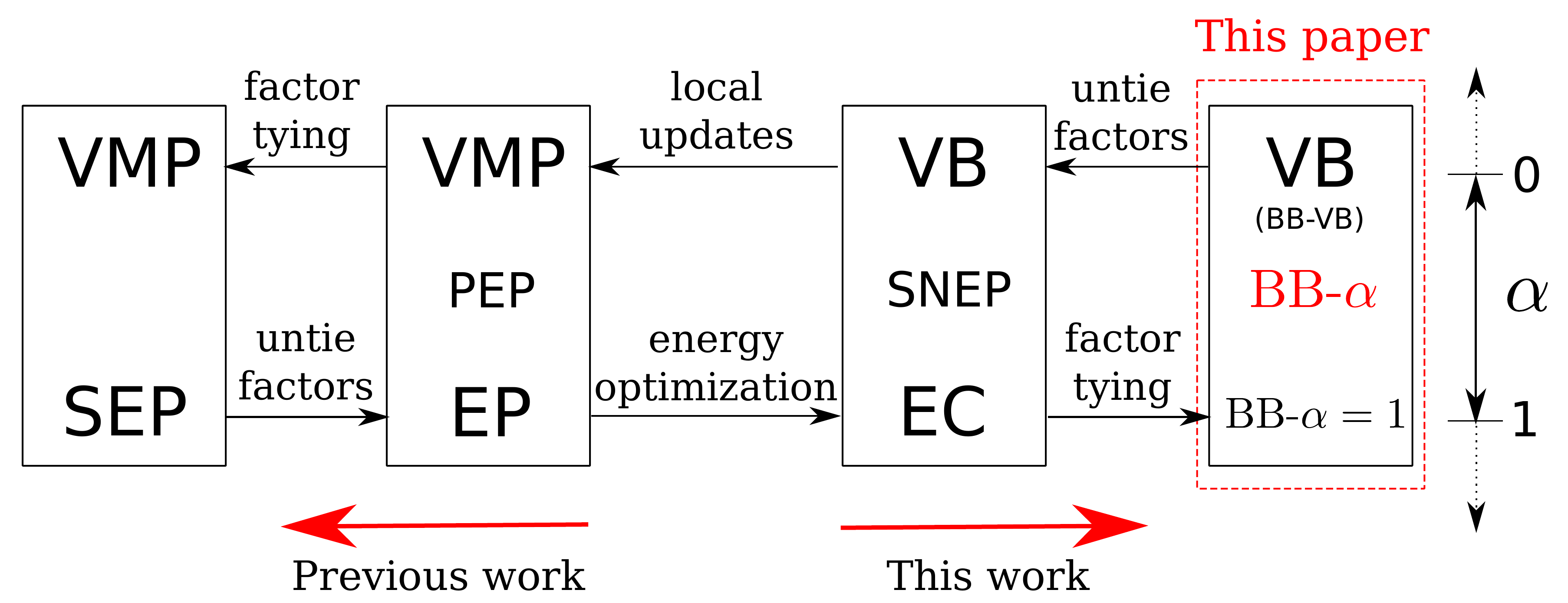}
\caption{Connections between the proposed black-box alpha (BB-$\alpha$) algorithm and a number of existing approaches. EC abbreviates expectation consistent approximate inference \cite{opper2005expectation} which is a double-loop convergent algorithm similar to \citet{heskes2002expectation}. VMP stands for variational message passing described in \citet{minka2005}, which should be distinguished from another version in \citet{winn2005}. Other acronyms are explained in the main text.}
\label{fig:relationship}
\vspace{-0.2in}
\end{figure}

%%%%%%%%%%% how BB-alpha differs from stochastic EP %%%%%%%%5
Buoyed by the success of 
%the %seemingly severe approximation of 
factor tying methods (SEP/AEP) , we % have
propose 
%evaluated a different approach in the same spirit that 
to apply the same idea directly to the power EP energy function, rather than
through the power EP message passing updates as in SEP/AEP. We call this new
method Black-box alpha (BB-$\alpha$). Figure
\ref{fig:relationship} illustrates its differences and
connections to other existing methods. Besides being memory efficient as SEP/AEP, BB-$\alpha$
has an analytic energy form that does not require double-loop procedures and
can be directly optimized using gradient descent. This means that popular
stochastic optimization methods can be used for large-scale learning with
BB-$\alpha$.

%%%%%%%%% black box inference for VB (EP) %%%%%%%%%%%%
An advantage of BB-$\alpha$, that gives origin to its name, is that it is a
black-box method that can be straightforward applied to very complicated
probabilistic models. In such models, the energy function of VB, EP and
power EP does not exist in an analytic form. \emph{Black-box} VB
\cite{ranganath2014black} sidesteps this difficulty by using Monte Carlo
approximations of the VB energy, see \cite{salimans2013fixed} for a related technique.
In this work, we follow a similar approach and
also approximate the BB-$\alpha$ energy by Monte Carlo. Similar approximations
have already been applied to EP \cite{barthelme2011abc, gelman2014,
xu2014,teh2015}. However, these methods do not use the factor tying idea and
consequently, are based on expensive double-loop approaches or on message
passing algorithms that lack convergence guarantees.

%In our paper, simple Monte
%Carlo is considered as a mean of approximating the energy function, although
%future work will also look at more sophisticated methods including MCMC and
%importance sampling. 
%
%Automatic differentiation techniques are also extended to the (approximate) EP context to allow fast model prototyping.

%
%We name our approach \emph{black-box alpha} (BB-$\alpha$), since it also extends to power EP which locally minimizes an $\alpha$-divergence. The proposed method can interpolate between VB ($\alpha \rightarrow 0$) and an EP-like method ($\alpha = 1$), and extrapolate beyond to $\alpha \rightarrow \pm \infty$. Empirical results on probit regression and neural network regression/classification problems demonstrate the scalability and accuracy of the proposed approximate (power) EP approach.

%% file: section2.tex
\section{$\alpha$-Divergence and power EP}
% alpha divergence part
Let us begin by briefly reviewing the $\alpha$-divergence upon which our method is based. 
Consider two probability densities $p$ and $q$ of a random variable $\bm{\theta}$; one fundamental question is to assess how close the two distributions are. 
The $\alpha$-divergence \cite{amari1985} measures the ``similarity'' between two distributions, and in this paper we adopt a more convenient form\footnote{Equivalent to the original definition by setting $\alpha' = 2 \alpha - 1$ in Amari's notation.} \cite{zhu1995}:
\begin{equation}
 \mathrm{D}_{\alpha}[p||q] = \frac{1}{\alpha (1 - \alpha)} \left( 1 - \int p(\bm{\theta})^{\alpha} q(\bm{\theta})^{1 - \alpha} d\bm{\theta}\right)\,.
\end{equation}
The following examples with different $\alpha$ values are interesting special cases:
\begin{equation}
  \mathrm{D}_{1}[p||q] = \lim_{\alpha \rightarrow 1}  \mathrm{D}_{\alpha}[p||q] =  \mathrm{KL}[p||q]\,,
\end{equation}
\begin{equation}
  \mathrm{D}_{0}[p||q] = \lim_{\alpha \rightarrow 0}  \mathrm{D}_{\alpha}[p||q] =  \mathrm{KL}[q||p]\,,
\end{equation}
{\small
\begin{equation}
  \mathrm{D}_{\frac{1}{2}}[p||q] = 2 \int \left(\sqrt{p(\bm{\theta})} - \sqrt{q(\bm{\theta})} \right)^2 d\bm{\theta} = 4\mathrm{Hel}^2[p||q]\,.
  \label{eq:hellinger}
\end{equation}
}For the first two limiting cases $\mathrm{KL}[p||q]$ denotes the
\emph{Kullback-Leibler (KL) divergence} given by $\mathrm{KL}[p||q] =
\mathbf{E}_{p}[\log p(\bm{\theta}) - \log q(\bm{\theta})]$. In
(\ref{eq:hellinger}) $\mathrm{Hel}[p||q]$ denotes the \emph{Hellinger distance}
between two distributions; $\mathrm{D}_{\frac{1}{2}}$ is the only 
member of the family of $\alpha$-divergences that is symmetric in $p$ and $q$.

\begin{figure}
\centering
 \includegraphics[width=1\linewidth]{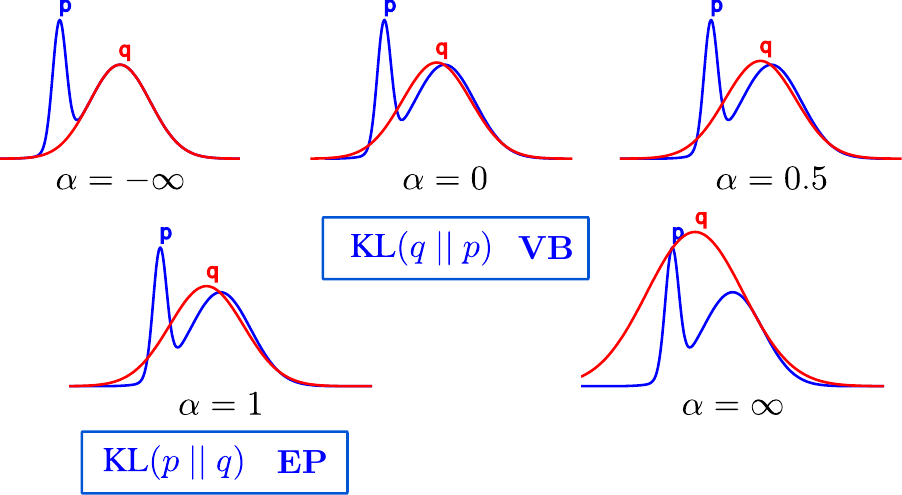}
 \caption{An illustration of approximating distributions by $\alpha$-divergence minimization. Here $p$ and $q$ shown in the graphs are unnormalized probability densities. Reproduced from \citet{minka2005}. Best viewed in color.}
 \label{fig:alpha_divergence}
\end{figure}

To understand how the choice of $\alpha$ might affect the result of approximate inference, consider the problem of approximating a complicated distribution $p$ with a tractable Gaussian distribution $q$ by minimizing $\mathrm{D}_{\alpha}[p||q]$. The resulting (unnormalized) approximations obtained for different values of $\alpha$ are visualized in Figure \ref{fig:alpha_divergence}. This shows that when $\alpha$ is a large positive number the approximation $q$ tends to cover all the modes of $p$, while for $\alpha \rightarrow -\infty$ (assuming the divergence is finite) $q$ is attracted to the mode with the largest probability mass. The optimal setting of $\alpha$ might reasonably be expected to depend on the learning task that is being considered.

Setting aside the analytic tractability of the computations, we note that the
minimization of a global $\alpha$-divergence might not always be desirable. If
the true posterior has many modes, a global approximation of this flavor that
is refined using $\alpha \geq 1$ will cover the modes, and can place
substantial probability in the area where the true posterior has low
probability (see the last plot in Figure \ref{fig:alpha_divergence}). 
The power EP algorithm \cite{minka2001, minka2004} minimizes instead a set of local
$\alpha$-divergences.
We now give a brief review of the power EP algorithm. Recall the definition of the typically intractable posterior distribution
\begin{equation} 
p(\bm{\theta}|\mathcal{D}) \propto \left[ \prod_{n=1}^N p(\bm{x}_n|\bm{\theta}) \right] p_0(\bm{\theta})\,.\label{eq:exact_posterior}
\end{equation}
Here for simplicity, the prior distribution $p_0(\bm{\theta}) = \exp\{\mathbf{s}(\bm{\theta})^{T} \bm{\lambda}_0 - \log Z(\bm{\lambda}_0) \}$ is assumed to have an exponential family form, where $\bm{\lambda}_0$ and $\mathbf{s}(\bm{\theta})$ are vectors of natural parameters and sufficient statistics, respectively, and $Z(\bm{\lambda}_0)$ is the normalization constant or partition function required to make $p_0(\bm{\theta})$ a valid distribution. 
We can use power EP to approximate the true posterior
$p(\bm{\theta}|\mathcal{D})$.  We define the site approximation
$f_n(\bm{\theta}) = \exp\{ \mathbf{s}(\bm{\theta})^{T} \bm{\lambda}_n \}$,
which is is within the same exponential family as the prior, and is used to
approximate the effect of the $n$-th likelihood factor $p(\bm{x}_n|\bm{\theta})$.
The approximate posterior is then defined as the product of all $f_n$ and the prior:
$q(\bm{\theta}) \propto \exp\{\mathbf{s}(\bm{\theta})^{T} (\sum_n
\bm{\lambda}_n + \bm{\lambda}_0) \}$.
In the
following sections we use $\bm{\lambda}_q$ to denote the natural parameters of
$q(\bm{\theta})$, and in the EP context we define $\bm{\lambda}_q = \sum_n \bm{\lambda}_n
+ \bm{\lambda}_0$. According to \citet*{minka2004} and \citet*{seeger2005}, the
power EP energy function with power $\alpha$ is
\begin{equation}
\textstyle
\begin{aligned}
E(\bm{\lambda}_0,\{\bm{\lambda}_n\}) =  \log Z(\bm{\lambda}_0) + \left(\frac{N}{\alpha}-1 \right) \log Z(\bm{\lambda}_q) \\
  - \frac{1}{\alpha} \sum_{n=1}^{N} \log \int p(\bm{x}_n|\bm{\theta})^{\alpha} \exp\{\mathbf{s}(\bm{\theta})^T (\bm{\lambda}_q - \alpha \bm{\lambda}_n )\} d\bm{\theta}\,.
\end{aligned}
\label{eq:energy}
\end{equation}
This energy is equal to minus the logarithm of the power EP approximation of the model evidence $p(\mathcal{D})$, that is, the normalizer of the right-hand side of (\ref{eq:exact_posterior}). Therefore, minimizing (\ref{eq:energy}) with respect to $\{\bm{\lambda}_n\}$ is arguably a sensible way to tune these variational parameters. However, power EP does not directly perform gradient descent to minimize $E(\bm{\lambda}_0,\{\bm{\lambda}_n\})$. Instead, it finds a stationary solution to the optimization problem by running a message passing algorithm that repeatedly applies the following four steps for every site approximation $f_n$:
\begin{itemize}
\item[1] Compute the cavity distribution: $$q^{\setminus n}(\bm{\theta}) \propto q(\bm{\theta}) / f_n(\bm{\theta})^{\alpha},$$ i.e. $\bm{\lambda}^{\setminus n} \bm{\leftarrow} \bm{\lambda}_q - \alpha \bm{\lambda}_n$;
\item[2] Compute the ``tilted'' distribution by inserting the exact likelihood factor raised to the power $\alpha$: $$\tilde{p}_n(\bm{\theta}) \propto q^{\setminus n}(\bm{\theta}) p(\bm{x}_n|\bm{\theta})^{\alpha};$$
\item[3] Adjust $q$ by matching moments: $$E_{q}[\bm{s}(\bm{\theta})] \leftarrow E_{\tilde{p}_n}[\bm{s}(\bm{\theta})];$$
\item[4] Recover the site approximation $f_n(\bm{\theta})$ by setting $\bm{\lambda}_n \leftarrow \bm{\lambda}_q - \bm{\lambda}^{\setminus n}$, and compute the final update for $q(\bm{\theta})$ by $\bm{\lambda}_q \leftarrow \sum_n \bm{\lambda}_n + \bm{\lambda}_0$.
\end{itemize}
Notice in step 3 moment matching is equivalent to updating the $q$ distribution by minimizing an $\alpha$-divergence, with the target proportional to $p(\bm{x}_n|\bm{\theta}) \exp\{\bm{s}(\bm{\theta})^T(\bm{\lambda}_q - \bm{\lambda}_n) \}$. To see this, consider approximating some $p$ distribution with $q$ by minimizing the $\alpha$-divergence; the gradient of $\mathrm{D}_{\alpha}[p||q]$ w.r.t.~the natural parameters $\bm{\lambda}_q$ is: 
\begin{equation}
 \begin{aligned}
 \nabla_{\bm{\lambda}_q} \mathrm{D}_{\alpha}[p||q] &= -\frac{1}{\alpha} \int p(\bm{\theta})^{\alpha} q(\bm{\theta})^{1 - \alpha} \nabla_{\bm{\lambda}_q} \log q(\bm{\theta}) d\bm{\theta} \\
 & = \frac{Z_{\tilde{p}}}{\alpha} \left( \mathbf{E}_{q}[\bm{s}(\bm{\theta})] - \mathbf{E}_{\tilde{p}}[\bm{s}(\bm{\theta})] \right)\,,
\end{aligned}
\label{eq:alpha_divergence_gradient}
\end{equation}
where $\tilde{p} \propto p^{\alpha} q^{1-\alpha}$ with normalization constant $Z_{\tilde{p}}$ and $\alpha \neq 0$. Substituting $p$ with $p' (\bm \theta) \propto p(\bm{x}_n|\bm{\theta}) \exp\{\bm{s}(\bm{\theta})^T(\bm{\lambda}_q - \bm{\lambda}_n) \}$ and zeroing the gradient results in step 3 that matches moments between the tilted distribution and the approximate posterior.
Also \citet*{minka2004} derived the stationary condition of (\ref{eq:energy}) as
\begin{equation}
\textstyle
\mathbf{E}_{\tilde{p}_n}[\mathbf{s}(\bm{\theta})] = \mathbf{E}_q[\mathbf{s}(\bm{\theta})],\  \forall n\,,
\label{eq:moment_matching_condition}
\end{equation}
so that it agrees with (\ref{eq:alpha_divergence_gradient}).
This means, at convergence, $q(\bm{\theta})$ minimizes the $\alpha$-divergences from all the tilted distributions to the approximate posterior.

Alternatively, \citet{heskes2002expectation} and \citet{opper2005expectation} proposed a convergent double-loop algorithm to solve the energy minimization problem for normal EP ($\alpha=1$) (see supplementary material). This algorithm first rewrites the energy (\ref{eq:energy}) as a function of the cavity parameters $\bm{\lambda}^{\setminus n}$ and adds the constraint $(N-1) \bm{\lambda}_q + \bm{\lambda}_0 = \sum_n \bm{\lambda}^{\setminus n}$ that ensures agreement between the global approximation and the local component approximate factors. It then alternates between an optimization of the cavity parameters in the inner loop and an optimization of the parameters of the posterior approximation $\bm{\lambda}_q$ in the outer loop. However, this alternating optimization procedure often requires too many iterations to be useful in practice.

%% file: section3.tex
\section{Approximate local minimization of $\alpha$-divergences}
\begin{figure}
 \centering
 \includegraphics[width=1\linewidth]{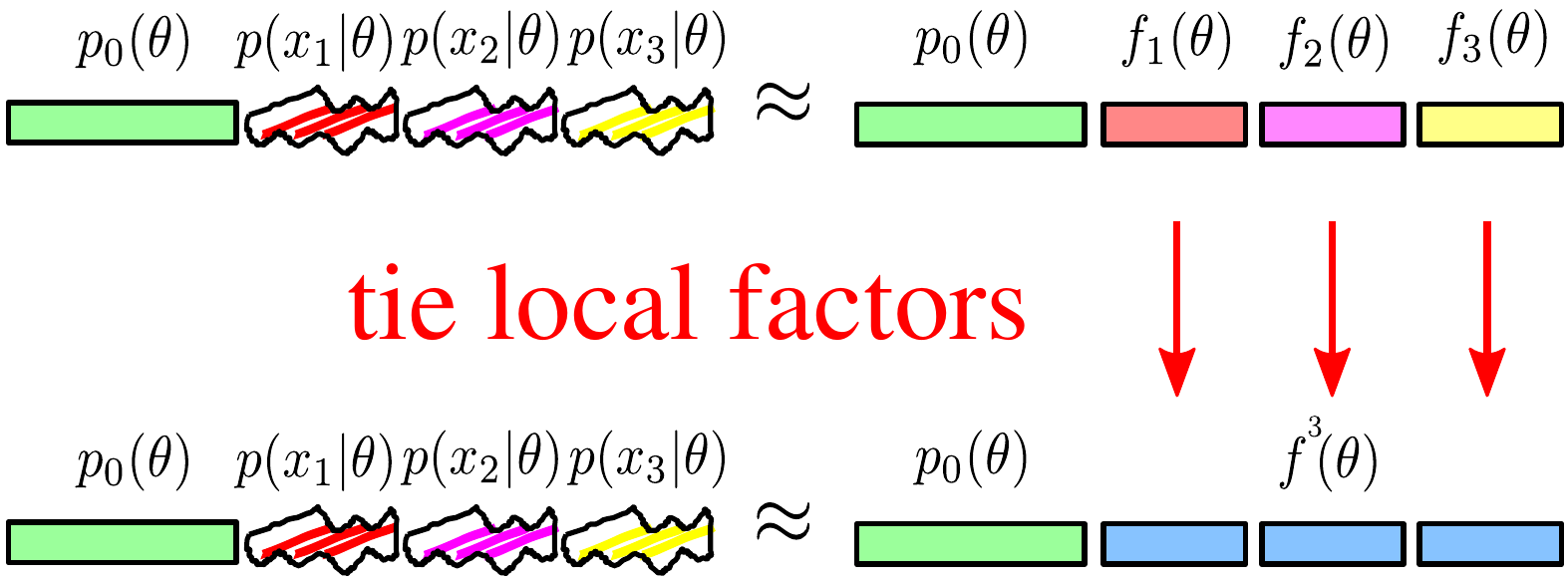}
 \caption{A cartoon for BB-$\alpha$'s factor tying constraint. Here we assume the dataset has $N=3$ observations. Best seen in color.}
 \label{fig:factor_tying}
 \vspace{-0.1in}
\end{figure}

In this section we introduce \emph{black-box alpha} (BB-$\alpha$), which approximates power EP with a simplified objective. 
Now we constrain all the site parameters to be equal, i.e.~$\bm{\lambda}_n = \bm{\lambda}$ for all $n$. This is equivalent to tying all the local factors, where now $f_n(\bm{\theta}) = f(\bm{\theta})$ for all $n$. Then all the cavity distributions are the same with natural parameter $\bm{\lambda}^{\setminus n} = (N - \alpha) \bm{\lambda} + \bm{\lambda}_0$, and the approximate posterior is parameterized by $\bm{\lambda}_q = N \bm{\lambda} + \bm{\lambda}_0$. Recall that $f_n(\bm{\theta})$ captures the contribution of the $n$-th likelihood to the posterior. Now with shared site parameters we are using an ``average site approximation'' $f(\bm{\theta}) = \exp\{\bm{s}(\bm{\theta})^T \bm{\lambda} \}$ that approximates the average effect of each likelihood term on the posterior. Under this assumption we rewrite the energy function (\ref{eq:energy}) by replacing $\bm{\lambda}_n$ with $\bm{\lambda}$:
\begin{equation}
\textstyle
\begin{aligned}
E(\bm{\lambda}_0, \bm{\lambda}) =& \log Z(\bm{\lambda}_0) - \log Z(\bm{\lambda}_q) \\
&- \frac{1}{\alpha} \sum_{n=1}^{N} \log \mathbf{E}_{q} \left[ \left( \frac{p(\bm{x}_n|\bm{\theta})}{f(\bm{\theta})} \right)^{\alpha} \right]\,.
\end{aligned}
\label{eq:bb_alpha_energy}
\end{equation}  
Figure \ref{fig:factor_tying} illustrates the comparison between the original power EP and the proposed method. Also, as there is a one-to-one correspondence between $\bm{\lambda}$ and $\bm{\lambda}_q$ given $\bm{\lambda}_0$, i.e.~$\bm{\lambda} = (\bm{\lambda}_q - \bm{\lambda}_0 )/N$, we can therefore rewrite (\ref{eq:bb_alpha_energy}) as $E(\bm{\lambda}_0, \bm{\lambda}_q)$ using the global parameter $\bm{\lambda}_q$. 

The factor tying constraint was proposed in \citet{li2015stochastic} and \citet{dehaene2015} to obtain versions of EP called \emph{stochastic EP} (SEP) and \emph{averaged EP} (AEP), respectively, thus the new method also inherits the advantage of memory efficiency.
However, applying this same idea directly to the energy function (\ref{eq:energy}) results in a different class of algorithms from AEP, SEP and power SEP. The main difference is that, when an exponential family approximation is considered, SEP averages the \emph{natural parameters} of the approximate posteriors obtained in step 3 of the message passing algorithm from the previous section. However, in BB-$\alpha$ the \emph{moments}, also called the \emph{mean parameters}, of the approximate posteriors are averaged and then converted to the corresponding natural parameters\footnote{Under a minimal exponential family assumption, there exists a one-to-one correspondence between the natural parameter and the mean parameter.}. To see this, one can take derivatives of $E(\bm{\lambda}_0, \bm{\lambda}_q)$ w.r.t.~$\bm{\lambda}_q$ and obtain the new stationary conditions for $\alpha \neq 0$ and $\alpha \neq N$:
\begin{align}
%\textstyle
\mathbf{E}_q[\mathbf{s}(\bm{\theta})]=\frac{1}{N}\sum_{n=1}^N\mathbf{E}_{\tilde{p}_n}[\mathbf{s}(\bm{\theta})]\,.
\label{eq:new_moment_matching_conditions}
\end{align}
Therefore, the moments of the $q$ distribution, i.e.~the expectation of $\mathbf{s}(\bm{\theta})$ with respect to $q(\bm{\theta})$, is equal to the average of the expectation of $\mathbf{s}(\bm{\theta})$ across the different tilted distributions $\tilde{p}_n(\bm{\theta}) \propto p(\bm{x}_n|\bm{\theta})^{\alpha} q^{\setminus n}(\bm{\theta})$, for $n = 1,\ldots,N$.  
It might not always be sensible to average moments, e.g.~this approach is shown
to be biased even in the simple case where the likelihood terms also belong to
the same exponential family as the prior and approximate posterior (see
supplementary material). Moreover, BB-$\alpha$ can mode-average, e.g. when
approximating the posterior distribution of component means in a Gaussian
Mixture Model, which may or may not be desired depending on the application.
But unlike SEP, the new method explicitly defines an energy function as an
optimization objective, which enables analysis of convergence and applications
of stochastic optimization/adaptive learning rate methods. Moreover, it can
provide an approximate posterior with better uncertainty estimates than VB (see
supplementary), which is desirable. The energy also makes hyper-parameter
optimization simple, which is key for many applications.

We prove the convergence of the new approximate EP method by showing that the new energy function (\ref{eq:bb_alpha_energy}) is bounded below for $\alpha \leq N$ when the energy function is finite. First, using Jensen's inequality, we can prove that the third term $- \frac{1}{\alpha} \sum_{n=1}^{N} \log \mathbf{E}_{q} \left[ \left( \frac{p(\bm{x}_n|\bm{\theta})}{f(\bm{\theta})} \right)^{\alpha} \right]$ in (\ref{eq:bb_alpha_energy}) is non-increasing in $\alpha$, because for arbitrary numbers $0< \alpha < \beta$ or $\beta < \alpha < 0$ the function $x^{\frac{\alpha}{\beta}}$ is strictly concave on $x \geq 0$. For convenience we shorthand this term as $G(\alpha)$. Then the proof is finished by subtracting from (\ref{eq:bb_alpha_energy}) $\tilde{G}(N) := - \frac{1}{N} \sum_n \log \int p_0(\bm{\theta}) p(\bm{x}_n|\bm{\theta})^N d\bm{\theta},$ 
a function that is independent of the $q$ distribution:
\begin{equation*}
 \begin{aligned}
  &E(\bm{\lambda}_0, \bm{\lambda}_q) - \tilde{G}(N) \\
  =& G(\alpha) + \log Z(\bm{\lambda}_0) - \log Z(\bm{\lambda}_q) - \tilde{G}(N) \\
  =& G(\alpha) - G(N) \geq 0\,.
 \end{aligned}
\end{equation*}

The stationary point of (\ref{eq:energy}) is expected to converge to the stationary point of (\ref{eq:bb_alpha_energy}) when more and more data are available. More precisely, as $N$ grows, we expect $q(\bm{\theta})$ and the cavities to become very concentrated. When this happens, the contribution of each likelihood factor $p(\bm{x}_n|\bm{\theta})$ to the tilted distribution $\tilde{p}_n(\bm{\theta})$ becomes very small because the likelihood is a rather flat function when compared to the cavity distribution $q^{\setminus n}(\bm{\theta})$. Therefore, as the amount of data $N$ increases, we expect all the terms $\mathbf{E}_{\tilde{p}_n}[\mathbf{s}(\bm{\theta})]$ in (\ref{eq:new_moment_matching_conditions}) to be very similar to each other. When all of them are equal, we have that (\ref{eq:new_moment_matching_conditions}) implies (\ref{eq:moment_matching_condition}).

As in power EP, which value of $\alpha$ returns the best approximation depends on the particular application. In the limit that $\alpha$ approaches zero, the BB-$\alpha$ energy (\ref{eq:bb_alpha_energy}) converges to the variational free energy:
\begin{equation*}
\begin{aligned}
 \textstyle
 & \lim_{\alpha \rightarrow 0} E(\bm{\lambda}_0, \bm{\lambda}_q) \\
 &= \log Z(\bm{\lambda}_0) - \log Z(\bm{\lambda}_q) - \sum_{n=1}^N \mathbf{E}_{q} \left[ \log \frac{p(\bm{x}_n|\bm{\theta})}{f(\bm{\theta})} \right] \\ 
 &= - \mathbf{E}_q \left[ \log \frac{\prod_n p(\bm{x}_n|\bm{\theta}) p_0(\bm{\theta})} {\exp \{ \bm{s}(\bm{\theta})^T \bm{\lambda}_q \} / Z(\bm{\lambda}_q) } \right] \\
 &= -\mathbf{E}_q \left[ \log p(\bm{\theta}, \mathcal{D}) - \log q(\bm{\theta}) \right] \\
 &= E_{VB}(\bm{\lambda}_0, \bm{\lambda}_q)\,.
\end{aligned}
\end{equation*}
Note also, the $\alpha$-divergence as a function of $\alpha$ is smooth in $[0, 1]$. Therefore, by adjusting the $\alpha$ parameter, we can interpolate between the solutions given by variational Bayes and the solutions given by the resulting approximation to EP. In the supplementary material we demonstrate this continuity in $\alpha$ with a linear regression example.

The prior hyper-parameters $\bm{\lambda}_0$ can be optimally adjusted by minimizing
(\ref{eq:bb_alpha_energy}). This can be done through a variational EM-like procedure. First we optimize the energy function w.r.t.~the $q$ distribution until convergence. Then we take the gradients w.r.t.~$\bm{\lambda}_0$ and equate them to zero to obtain the update for the prior, which is
$\mathbf{E}_q[\mathbf{s}(\bm{\theta})] = \mathbf{E}_{p_0}[\mathbf{s}(\bm{\theta})].$ However, this procedure is inefficient in practice. Instead we jointly optimize the approximate posterior $q$ and the prior distribution $p_0$ similar to the approach of \citet{hernandez2016scalable} for normal EP.

\subsection{Large scale learning}

When $N$ is large, it might be beneficial to minimize (\ref{eq:bb_alpha_energy}) using stochastic optimization techniques. In particular, we can uniformly sample a mini-batch of data $\mathbf{S}\subseteq \{1,\ldots,N\}$ and construct the noisy estimate of the energy function given by
\begin{equation}
\begin{aligned}
\textstyle
E(\bm{\lambda}_0, \bm{\lambda}_q) &\approx \log Z(\bm{\lambda}_0) - \log Z(\bm{\lambda}_q) \\ &- \frac{1}{\alpha} \frac{N}{|\mathbf{S}|} \sum_{n \in \mathbf{S}} \log \mathbf{E}_{q} \left[ \left( \frac{p(\bm{x}_n|\bm{\theta})}{f(\bm{\theta})} \right)^{\alpha} \right]\,.
\end{aligned}
\label{eq:noisy_alpha_energy}
\end{equation}
The gradients of (\ref{eq:noisy_alpha_energy}) can then be used to minimize the
original objective by stochastic gradient descent. Under mild conditions, as discussed by \citet{bottou1998}, and using a learning rate schedule $\{\gamma_t\}$ that satisfies the Robbins-Monro conditions
$$ \sum_{t=1}^{\infty} \gamma_t = \infty, \quad \sum_{t=1}^{\infty} \gamma_t^2 < \infty,$$
the stochastic optimization procedure will converge to a fixed point of the exact energy function (\ref{eq:bb_alpha_energy}).

Similar to SEP/AEP \cite{li2015stochastic, dehaene2015}, BB-$\alpha$ only maintains the global parameter $\bm{\lambda}_q$ and the prior parameter $\bm{\lambda}_0$. This has been shown to achieve a significant amount of memory saving. On the other hand, recent work on parallelizing EP \cite{gelman2014, xu2014, teh2015}, whether in a synchronous manner or not, extends to BB-$\alpha$ naturally. But unlike EP, which computes different cavity distributions for different data points, BB-$\alpha$ uses the same cavity distribution for each data point. 

%%%%%%%%%%%%
\subsection{Black-box $\alpha$-divergence minimization}\label{sec:black_box_alpha}
In complicated probabilistic models, we might not be able to analytically compute the expectation over the approximate distribution $\mathbf{E}_{q} \left[ \left(\frac{p(\bm{x}_n|\bm{\theta})}{f(\bm{\theta})}  \right)^{\alpha} \right]$ in (\ref{eq:noisy_alpha_energy}) involving the likelihood
factors.  However, we can obtain an estimate of these integrals by Monte Carlo.
In this work we use the simplest method for doing this, but techniques like SMC and MCMC could also have the potential to be deployed.
We draw $K$ samples $\bm{\theta}_1,\ldots,\bm{\theta}_K$ from
$q(\bm{\theta})$ and then approximate the integrals by expectations with respect to those samples. This produces the following noisy estimate of the energy function:
\begin{equation}
 \begin{aligned}
\textstyle
\hat{E}(\bm{\lambda}_0, \bm{\lambda}_q) &= \log Z(\bm{\lambda}_0) - \log Z(\bm{\lambda}_q) \\ &- \frac{1}{\alpha} \frac{N}{|\mathbf{S}|} \sum_{n \in \mathbf{S}} \log \frac{1}{K} \sum_k \left( \frac{p(\bm{x}_n|\bm{\theta}_k)}{f(\bm{\theta}_k)} \right)^{\alpha}\,.
\end{aligned}
\label{eq:new_noisy_alpha_energy}
\end{equation}
Note, however, that the resulting
stochastic gradients will be biased because the energy function
(\ref{eq:new_noisy_alpha_energy}) applies a non-linear transformation (the
logarithm) to the Monte Carlo estimator of the integrals. Nevertheless, this
bias can be reduced by increasing the number of samples $K$. Our experiments indicate that when $K \geq 10$ the bias is almost negligible w.r.t.~the variance from sub-sampling the data using minibatches, for the models considered here.

There are two other tricks we have used in our implementation. The first one is the \emph{reparameterization trick} \cite{Kingma2014}, which has been used to reduce the variance of the Monte Carlo approximation to the variational free energy. Consider the case of computing expectation $\mathbf{E}_{q(\bm{\theta})}[F(\bm{\theta})]$. This expectation can also be computed as $\mathbf{E}_{p(\bm{\epsilon})}[F(g(\bm{\epsilon}))]$, if there exists a mapping $g(\cdot)$ and a distribution $p(\bm{\epsilon})$ such that $\bm{\theta} = g(\bm{\epsilon})$ and $q(\bm{\theta})d\bm{\theta} = p(\bm{\epsilon}) d\bm{\epsilon}$. Now consider a Gaussian approximation $q(\bm{\theta})$ as a running example. Since the Gaussian distribution also has an (minimal) exponential family form, there exists a one-to-one correspondence between the mean parameters $\{\bm{\mu}, \bm{\Sigma}\}$ and the natural parameters $\bm{\lambda}_q = \{\bm{\Sigma}^{-1} \bm{\mu}, \bm{\Sigma}^{-1} \}$. Furthermore, sampling $\bm{\theta} \sim q(\bm{\theta})$ is equivalent to $\bm{\theta} = g(\bm{\epsilon}) = \bm{\mu} + \bm{\Sigma}^{1/2} \bm{\epsilon}, \bm{\epsilon} \sim \mathcal{N}(\bm{0}, \bm{I})$. Thus the sampling approximation $\hat{E}(\bm{\lambda}_0, \bm{\lambda}_q)$ can be computed as
\begin{equation}
 \begin{aligned}
\textstyle
\hat{E}(\bm{\lambda}_0, \bm{\lambda}_q) &= \log Z(\bm{\lambda}_0) - \log Z(\bm{\lambda}_q) \\ &- \frac{1}{\alpha} \frac{N}{|\mathbf{S}|} \sum_{n \in \mathbf{S}} \log \frac{1}{K} \sum_k \left( \frac{p(\bm{x}_n|g(\bm{\epsilon}_k))}{f(g(\bm{\epsilon}_k))} \right)^{\alpha}\,,
\end{aligned}
\label{eq:new_noisy_alpha_energy_reparam}
\end{equation}
with $\bm{\epsilon}_1, ..., \bm{\epsilon}_K$ sampled from a zero mean, unit variance Gaussian. A further trick is resampling $\{\bm{\epsilon}_k\}$ every $M > 1$ iterations.  
In our experiments with neural networks, this speeds-up the training process since it reduces the transference of randomness to the GPU, which slows down computations.

Given a new probabilistic model, one can then use the proposed approach to
quickly implement, in an automatic manner, an inference algorithm based on the
local minimization of $\alpha$-divergences. For this one only needs to write
code that evaluates the likelihood factors $f_1,\ldots,f_N$ in
(\ref{eq:new_noisy_alpha_energy_reparam}). After this, the most difficult task is the
computation of the gradients of (\ref{eq:new_noisy_alpha_energy_reparam}) so that
stochastic gradient descent with minibatches can be used to optimize the energy
function.  However, the computation of these gradients can be easily automated
by using automatic differentiation tools such as Autograd
(\url{http://github.com/HIPS/autograd}) or Theano \cite{Bastien-Theano-2012}.
This approach allows us to quickly implement and test different modeling assumptions with little effort.

%% file: experiments.tex
\section{Experiments}

We evaluated the proposed algorithm black-box alpha (BB-$\alpha$),
on regression and classification problems using a probit regression model and 
Bayesian neural networks. The code for BB-$\alpha$ is publicly available\footnote{\url{https://bitbucket.org/jmh233/code_black_box_alpha_icml_2016}}
We also compare with a method that optimizes a Monte
Carlo approximation to the variational lower bound \cite{ranganath2014black}.
This approximation is obtained in a similar way to the one described in Section
\ref{sec:black_box_alpha}, where one can show its equivalence to BB-$\alpha$ by
limiting $\alpha \rightarrow 0$. We call this method black-box variational Bayes
(BB-VB). In the implementation of BB-$\alpha$ and BB-VB shown here, 
the posterior approximation $q$ is always a
factorized Gaussian (but more complex distributions can easily be handled). 
The mean parameters of $q$ are initialized by
independently sampling from a zero mean Gaussian with standard deviation
$10^{-1}$. We optimize the logarithm of the variance parameters in $q$ to avoid
obtaining negative variances. We use -10 as the initial value for the
log-variance parameters, which results in very low initial variance in $q$.
This makes the stochastic optimizer initially resemble a point estimator method
which quickly finds a solution for the mean parameters with good predictive
properties on the training data. After this, the stochastic optimizer
progressively increases the variance parameters to capture the uncertainty
around the mean of $q$. This trick considerably improves the performance of the
stochastic optimization method for BB-$\alpha$ and BB-VB. The prior
$p(\mathbf{x})$ is always taken to be a factorized Gaussian with zero mean and unit
standard deviation. The implementation of each of the analyzed methods is
available for reproducible research.

\input{tables/table_probit_all}

\subsection{Probit regression}

We perform experiments with a Bayesian probit regression model to validate the
proposed black-box approach. We optimize the different objective functions
using minibatches of size 32 and Adam \cite{kingma2014adam} with its default
parameter values during 200 epochs. BB-$\alpha$ and BB-VB are implemented by
drawing $K=100$ Monte Carlo samples for each minibatch. The following values
$\alpha=1$, $\alpha=0.5$ and $\alpha=10^{-6}$ for BB-$\alpha$ are considered 
in the experiments. The performance of each method is evaluated on 50 random 
training and test splits of the data with 90\% and 10\% of the data instances, 
respectively.  

Table \ref{tab:probit_results} shows the average test log-likelihood and
test error obtained by each technique in the probit regression datasets 
from the UCI data repository \cite{Lichman:2013}.
We also show the average rank obtained by each method across all the train/test
splits of the data. Overall, all the methods obtain very similar results
although BB-$\alpha$ with $\alpha=1.0$ seems to perform slightly better. 
Importantly, BB-$\alpha$ with $\alpha=10^{-6}$ produces the same results as
BB-VB, which verifies our theory of continuous interpolations near the limit
$\alpha \rightarrow 0$.

\subsection{Neural network regression}\label{sec:experiments_neural_network_regression}

We perform additional experiments with neural networks for regression with 100
units in a single hidden layer with ReLUs and Gaussian additive noise at the
output. We consider several regression datasets also from the UCI data repository
\cite{Lichman:2013}. We use the same training procedure as before, but using
500 epochs instead of 200. The noise variance is learned in each method by
optimizing the corresponding objective function: evidence lower bound in BB-VB
and (\ref{eq:bb_alpha_energy}) in BB-$\alpha$.

\begin{figure}
\center
\includegraphics[width=1.0\linewidth]{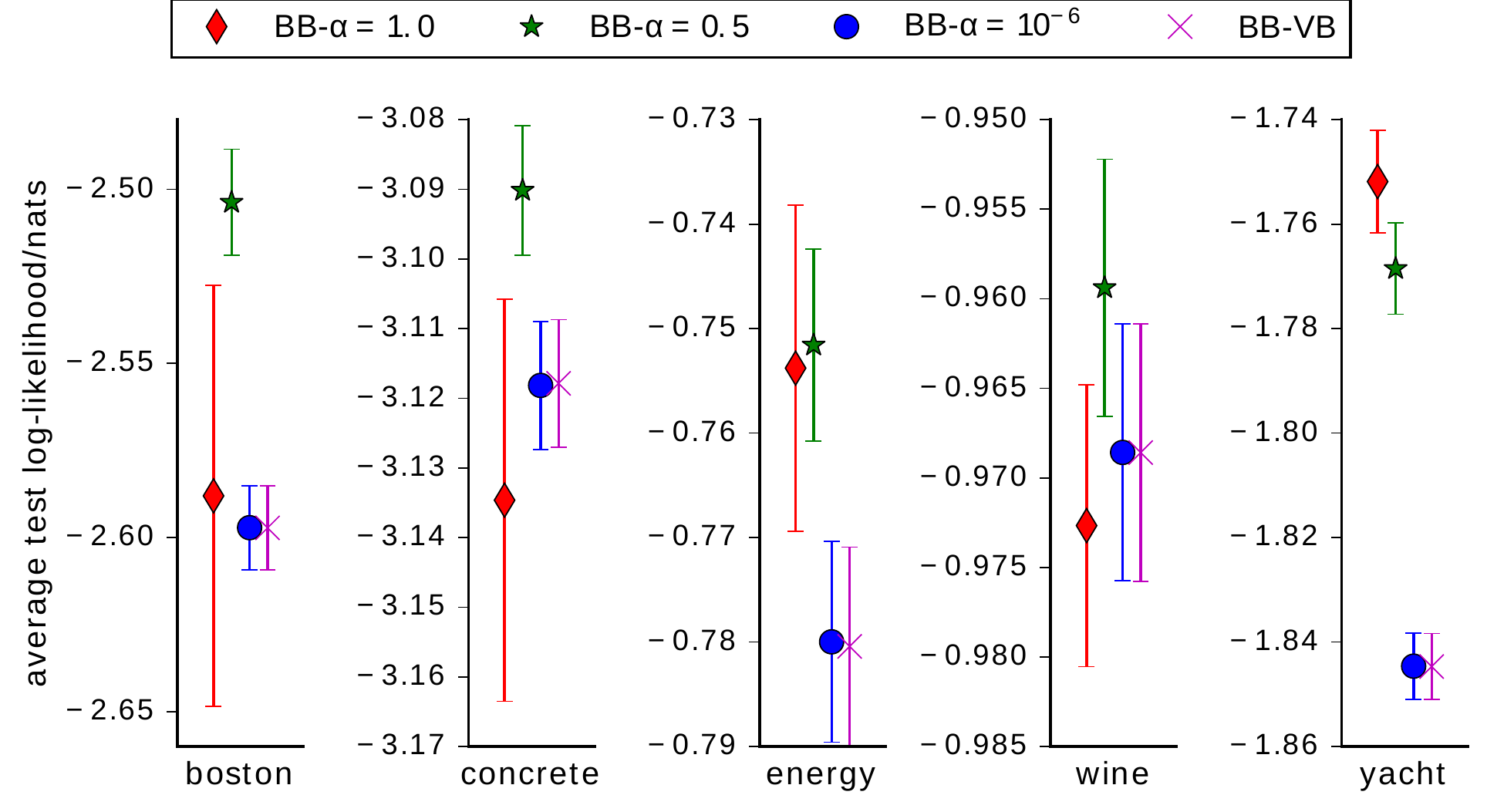}
\caption{Average test log-likelihood and the ranking comparisons. Best viewed in color.}
\label{fig:regression_ll}
\end{figure}

The average test log-likelihood across 50 splits of the data into training
and test sets are visualized in Figure \ref{fig:regression_ll}. The performance
of BB-$\alpha=10^{-6}$ is again almost indistinguishable from that of BB-VB.
None of the tested $\alpha$ settings clearly dominates the other choices in all
cases, indicating that the optimal $\alpha$ may vary for different
problems.  However, $\alpha=0.5$ produces good overall results. This might be
because the Hellinger distance is the only symmetric divergence measure in the
$\alpha$-divergence family which balances the tendencies of capturing a mode
($\alpha < 0.5$) and covering the whole probability mass ($\alpha > 0.5$). There
are no significant differences in regression error so the results are not shown.

\subsection{Neural network classification}

The next experiment considers the MNIST digit classification problem. We use neural
networks with 2 hidden layers with 400 hidden units per layer, ReLUs and
softmax outputs. In this problem, we initialize the mean parameters in $q$ as
recommended in \citet{glorot2010understanding} by sampling from a zero-mean
Gaussian with variance given by $2 / ($d$_\text{in} + $d$_\text{out})$, where
d$_\text{in}$ and d$_\text{out}$ are the dimensionalities of the previous and
the next layer in the neural network.  In this case we also try $\alpha=-1.0$
in BB-$\alpha$. We use minibatches of size 250 and run the different methods
for 250 epochs.  BB-$\alpha$ and BB-VB use now $K=50$ Monte Carlo samples to
approximate the expectations with respect to $q$ in each minibatch.  We
implemented the neural networks in Theano and ran the different methods on
GPUs using Adam with its default parameter values and a learning rate of
$0.0001$. In this case, to reduce the transference of data from the main memory 
to the GPU, we update the randomness in the Monte Carlo samples only after 
processing 10 minibatches instead of after processing each minibatch.

\input{tables/table_mnist_all}
\input{tables/cep_table}

Table \ref{tab:mnist_results} summarizes the average test error and test
log-likelihood obtained by each method over 20 random initializations. For
non-negative alpha settings BB-VB ($\alpha=0$) returns the best result, and
again BB-$\alpha=10^{-6}$ performs almost identically to the variational
inference approach BB-VB.  The best performing method is BB-$\alpha$ with $\alpha =
-1$, which is expected to move slightly towards fitting a mode. We note here
that this is different from the Laplace approximation, which, depending on the
curvature at the MAP solution, might return an approximate posterior that also
covers spaces other than the mode. On this dataset $\alpha=-1$ returns both
higher test log-likelihood and lower test error than all the tested
non-negative settings for $\alpha$. 

\subsection{Clean energy project data}

We perform additional experiments with data from the Harvard Clean Energy
Project, which is the world's largest materials high-throughput virtual
screening effort \cite{hachmann2014lead}. It has scanned a large number of
molecules of organic photovoltaics to find those with high power conversion
efficiency (PCE) using quantum-chemical techniques. The target value within
this dataset is the PCE of each molecule.  The input features for
all molecules in the data set are 512-bit Morgan circular fingerprints,
calculated with a bond radius of 2, and derived from the canonical smiles,
implemented in the RDkit. We use 50,000 molecules for training and 10,000
molecules for testing.

We consider neural networks with 2 hidden layers with 400 hidden units per
layer, ReLUs and Gaussian additive noise at the output. The noise variance is
fixed to 0.16, which is the optimal value according to the results reported in
\citet{pyzer2015learning}. The initialization and training process is the same
as with the MNIST dataset, but using the default learning rate in Adam.

Table \ref{tab:results_cep} summarizes the average test error and test
log-likelihood obtained by each method over 20 random initializations.
BB-$\alpha=10^{-6}$ obtains again very similar results to BB-VB.  In this case,
the best performing method is BB-$\alpha$ with $\alpha = 0.5$, both in terms of
test error and test log-likelihood. This was also the best performing method in
the experiments from Section \ref{sec:experiments_neural_network_regression},
which indicates that $\alpha=0.5$ may be a generally good setting in neural
network regression problems. Table \ref{tab:results_cep} shows that
$\alpha=0.5$ attains a balance between the tendency of $\alpha=1.0$ to perform
well in terms of test log-likelihood and the tendency of $\alpha=10^{-6}$ to
perform well in terms of test squared error.

\subsection{Analysis of the bias and variance in the gradients}
We perform another series of experiments to analyze the bias and the variance
in the gradients of (\ref{eq:new_noisy_alpha_energy_reparam}) as a function of
the number $K$ of Monte Carlo samples used to obtain a noisy approximation of
the objective function and the value of $\alpha$ used. 
To estimate the bias in
BB-$\alpha$ we run the Adam optimizer for 100 epochs on the Boston housing
dataset as described in Section \ref{sec:experiments_neural_network_regression}.
Then, we estimate the biased gradient using $K=1, 5, 10$ Monte Carlo samples from $q$, 
which is repeated 1,000 times to obtain an averaged estimate. We also compute an approximation to the ground truth for the unbiased gradient by using $K=10,000$ 
Monte Carlo samples. The whole process is performed across 15 minibatches of 
data points from each split.
We then define the bias in the gradient as the averaged L2 norm between the ground
truth gradient and the biased gradient across these 15 minibatches, divided by
the square root of the dimension of the gradient vector. This definition of
bias is not 0 for methods that use unbiased estimators of the gradient, such as
BB-VB, because of the variance by sampling on each minibatch.  However, we expect this procedure to
report larger bias values for BB-$\alpha$ than for BB-VB. Therefore, we
subtract from the bias values obtained for BB-$\alpha$ the corresponding
bias values obtained for BB-VB. This eliminates from the bias values any
additional variance that is produced by having to sample to estimate the
unbiased and biased gradient on each minibatch.

\input{tables/table_bias}
\input{tables/table_variance}

Table \ref{tab:results_bias} shows the average bias obtained for each value of
$K$ and $\alpha$ across the 50 splits.
We observe that the bias is reduced as we increase $K$ and as we make $\alpha$
closer to zero. For $K=10$ the bias is already very low. To put these bias
numbers into context, we also computed a standard deviation-like measure by
the square root of the average empirical variance per dimension in the noisy
gradient vector over the 15 minibatches. Table \ref{tab:results_variance} shows
the average values obtained across the 50 splits, where entries are almost constant
as a function of $\alpha$, and up to 5 orders of magnitude larger than
the entries of Table \ref{tab:results_bias} for $K=10$.  This means
the bias in the gradient used by BB-$\alpha$ is negligible when compared
with the variability that is obtained in the gradient by subsampling the
training data.

%% file: tables/table_probit_all.tex
\begin{table*}[!ht]
\centering
\caption{Probit regression experiment results}
\label{tab:probit_results}
\resizebox{\textwidth}{!}{%
\begin{tabular}{l@{\ica}|r@{$\pm$}l@{\ica}r@{$\pm$}l@{\ica}r@{$\pm$}l@{\ica}r@{$\pm$}l@{\ica}|r@{$\pm$}l@{\ica}r@{$\pm$}l@{\ica}r@{$\pm$}l@{\ica}r@{$\pm$}l@{\ica}}
\hline
&\multicolumn{8}{c|}{Average Test Log-likelihood}&\multicolumn{8}{c}{Average Test Error}\\
\bf{Dataset}&\multicolumn{2}{c}{\bf{BB-$\alpha{=}1.0$}}&\multicolumn{2}{c}{\bf{BB-$\alpha{=}0.5$}}&\multicolumn{2}{c}{\bf{BB-$\alpha{=}10^{-6}$}}&\multicolumn{2}{c|}{\bf{BB-VB}}&\multicolumn{2}{c}{\bf{BB-$\alpha{=}1.0$}}&\multicolumn{2}{c}{\bf{BB-$\alpha{=}0.5$}}&\multicolumn{2}{c}{\bf{BB-$\alpha{=}10^{-6}$}}&\multicolumn{2}{c}{\bf{BB-VB}}\\
\hline
Ionosphere&-0.333&0.022&-0.333&0.022&-0.333&0.022&\textbf{-0.333}&\textbf{0.022}&0.124&0.008&0.124&0.008&\textbf{0.123}&\textbf{0.008}&0.123&0.008\\
Madelon&\textbf{-0.799}&\textbf{0.006}&-0.920&0.008&-0.953&0.009&-0.953&0.009&\textbf{0.445}&\textbf{0.005}&0.454&0.004&0.457&0.005&0.457&0.005\\
Pima&\textbf{-0.501}&\textbf{0.010}&-0.501&0.010&-0.501&0.010&-0.501&0.010&\textbf{0.234}&\textbf{0.006}&0.234&0.006&0.235&0.006&0.235&0.006\\
Colon Cancer&\textbf{-2.261}&\textbf{0.402}&-2.264&0.403&-2.268&0.404&-2.268&0.404&\textbf{0.303}&\textbf{0.028}&0.307&0.028&0.307&0.028&0.307&0.028\\
\hline
\textbf{Avg. Rank}&\textbf{1.895}&\textbf{0.097}&2.290&0.038&2.970&0.073&2.845&0.072&\textbf{2.322}&\textbf{0.048}&2.513&0.039&2.587&0.031&2.578&0.031\\
\hline
\end{tabular}%
}
\end{table*}

%% file: tables/table_mnist_all.tex
\begin{table}[!ht]
\centering
\caption{Average Test Error and Log-likelihood in MNIST}
\label{tab:mnist_results}
\resizebox{0.85\columnwidth}{!}{%
\begin{tabular}{l@{\ica}|c@{\ica}c@{\ica}|c@{\ica}c}
\hline
Setting&Error/100&Rank&LL/100&Rank\\
\hline
BB-$\alpha=1.0$&1.51&4.97&-5.51&5.00\\
BB-$\alpha=0.5$&1.44&4.03&-5.09&4.00\\
BB-$\alpha=10^{-6}$&1.36&2.15&-4.68&2.55\\
BB-VB&1.36&2.12&-4.68&2.45\\
BB-$\alpha=-1.0$&\textbf{1.33}&\textbf{1.73}&\textbf{-4.47}&\textbf{1.00}\\
\hline
\end{tabular}%
}
\end{table}

%% file: tables/cep_table.tex
\begin{table*}[!ht]
\centering
\caption{Average Test Error and Test Log-likelihood in CEP Dataset.}
\begin{tabular}{l@{\ica}r@{$\pm$}l@{\ica}r@{$\pm$}l@{\ica}r@{$\pm$}l@{\ica}r@{$\pm$}l@{\ica}}\hline
\bf{CEP Dataset}&\multicolumn{2}{c}{\bf{ BB-$\alpha{=}1.0$ }}&\multicolumn{2}{c}{\bf{ BB-$\alpha{=}0.5$ }}&\multicolumn{2}{c}{\bf{ BB-$\alpha{=}10^{-6}$ }}&\multicolumn{2}{c}{\bf{ BB-VB }}\\
\hline
{\bf Avg. Error } & 1.28&0.01&\bf{1.08}&\bf{0.01}&1.13&0.01&1.14&0.01\\
{\bf Avg. Rank } &  4.00  &  0.00 & {\bf  1.35}  &  {\bf 0.15} &  2.05  &  0.15 &  2.60  &  0.13\\
\hline
{\bf Avg. Log-likelihood } &-0.93&0.01&\bf{-0.74}&\bf{0.01}&-1.39&0.03&-1.38&0.02\\
{\bf Avg. Rank } &  1.95  &  0.05 &  {\bf 1.05}  &  {\bf 0.05 } &  3.40  &  0.11 &  3.60  &  0.11\\
\hline
\end{tabular}
\label{tab:results_cep}
\end{table*}

%% file: tables/table_bias.tex
\begin{table}[t]
\centering
\caption{ Average Bias.}
\resizebox{0.97\columnwidth}{!}{
\begin{tabular}{c@{\ica}c@{\ica}c@{\ica}c@{\ica}}\hline
\bf{Dataset}&\bf{ BB-$\alpha{=}1.0$ }&\bf{ BB-$\alpha{=}0.5$ }&\bf{ BB-$\alpha{=}10^{-6}$ }\\
\hline
$K = 1$&0.2774&0.1214&0.0460\\
$K = 5$&0.0332&0.0189&0.0162\\
$K = 10$&0.0077&0.0013&0.0001\\
\hline
\end{tabular}
\label{tab:results_bias}
}
\end{table}

%% file: tables/table_variance.tex
\begin{table}[t]
\centering
\caption{ Average Standard Deviation Gradient. }
\resizebox{0.95\columnwidth}{!}{
\begin{tabular}{c@{\ica}c@{\ica}c@{\ica}c@{\ica}c@{\ica}}\hline
\bf{Dataset}&\bf{ BB-$\alpha{=}1.0$ }&\bf{ BB-$\alpha{=}0.5$ }&\bf{ BB-$\alpha{=}10^{-6}$ }\\
\hline
$K = 1$&14.1209&14.0159&13.9109\\
$K = 5$&12.7953&12.8418&12.8984\\
$K = 10$&12.3203&12.4101&12.5058\\
\hline
\end{tabular}
\label{tab:results_variance}
}
\end{table}

%% file: conclusion.tex
\section{Conclusions and future work}
We have proposed BB-$\alpha$ as a black-box inference algorithm to approximate
power EP. This is done by considering the energy function 
used by power EP and constraining the form of the site approximations in this
method. The proposed method locally minimizes the $\alpha$-divergence that is a rich
family of divergence measures between distributions including the Kullback-Leibler 
divergence. Such a method is guaranteed to converge under certain conditions, and can be implemented by optimizing an 
energy function without having to use inefficient double-loop algorithms. 
Scalability to large datasets can be achieved by using stochastic gradient
descent with minibatches. Furthermore, a combination of a Monte Carlo
approximation and automatic differentiation methods allows our technique to be
applied in a straightforward manner to a wide range of probabilistic models with
complex likelihood factors. Experiments with neural
networks applied to small and large datasets demonstrate both the accuracy and
the scalability of the proposed approach. The evaluations also indicate the optimal 
setting for $\alpha$ may vary for different tasks. Future work should provide a theoretical guide
and/or automatic tools for modelling with different factors and different $\alpha$ values. 

%% file: appendix.tex
%\documentclass[]{article}
%
%\usepackage{amsmath}
%\usepackage{amssymb}
%\usepackage{graphicx}
%\usepackage{bm}
%
%\usepackage{geometry}
%\geometry{margin=1in}
%
%
%% SPECIFIC PACKAGES USED IN THIS DOCUMENT
%\usepackage{wrapfig}
%\newcommand{\bt}{\boldsymbol{\theta}}
%\newcommand{\bx}{\mathbf{x}}
%%---------------------------------------
%
%
%\usepackage{parskip}
%\usepackage{authblk}
%\usepackage{hyperref}
%\setlength{\parindent}{0pt}
%
%%opening
%\title{Black-Box $\alpha$-Divergence Minimization: Supplementary}
%
%\author{Jos\'e Miguel Hern\'andez-Lobato$^{1*}$ \texttt{jmh@seas.harvard.edu}
%\and
%Yingzhen Li$^{2*}$ \texttt{yl494@cam.ac.uk}
%\and Mark Rowland$^2$ \texttt{mr504@cam.ac.uk}
%\and Daniel Hern\'andez-Lobato$^3$ \texttt{daniel.hernandez@uam.es}
%\and Richard E.~Turner$^2$ \texttt{ret26@cam.ac.uk}
%\and $^1$Harvard University, $^2$University of Cambridge, $^3$Universidad Aut\'onoma de Madrid,\\ $^*$Both authors contributed equally.
%}
%
%\date{}
%
%\begin{document}
%
%\maketitle

\appendix

\section{The Min-Max Problem of EP}
This section revisits the original EP algorithm as a min-max optimization problem.
Recall in the main text that we approximate the true posterior distribution $p(\bm{\theta}|\mathcal{D})$ with a distribution in exponential family form given by $q(\bm{\theta}) \propto \exp\{\mathbf{s}(\bm{\theta})^T \bm{\lambda}_q\}$. Now we define a set of \emph{unnormalized} cavity distributions $q^{\setminus n}(\bm{\theta}) = \exp\{\mathbf{s}(\bm{\theta})^T \bm{\lambda}_{\setminus n}\}$ for every data point $\bm{x}_n$. 
Then according to \cite{minka2001}, the EP energy function is
\begin{align}
\textstyle
E(\bm{\lambda}_q, \{ \bm{\lambda}_{\setminus n} \}) & =\textstyle \log Z(\bm{\lambda}_0) + ( N -1 )\log Z(\bm{\lambda}_q)
- \sum_{n=1}^{N}\log\int p(\bm{x}_n|\bm{\theta}) q^{\setminus n}(\bm{\theta}) d\bm{\theta} \,.\label{eq:energy}
\end{align}
In practice EP finds a stationary solution to the constrained optimization problem 
\begin{align}
\textstyle
\min_{\bm{\lambda}_q}\,\max_{\{\bm{\lambda}_n\}}\, E(\bm{\lambda},\{\bm{\lambda}_n\})
\quad
\textstyle
\text{subject to}\quad (N-1)\bm{\lambda}_q + \bm{\lambda}_0 =\sum_{n=1}^{N}\bm{\lambda}_{\setminus n}\,,
\label{eq:problem}
\end{align}
where the constraint in (\ref{eq:problem}) guarantees that the $\{\bm{\lambda}_{\setminus n} \}$ are valid cavity parameters that are consistent with the approximate posterior. 
Similarly for power EP the energy function has the following form:
\begin{align}
\textstyle
E(\bm{\lambda}_q, \{ \bm{\lambda}_{\setminus n} \}) & =\textstyle \log Z(\bm{\lambda}_0) + \left( \frac{N}{\alpha} -1 \right)\log Z(\bm{\lambda}_q)
- \frac{1}{\alpha} \sum_{n=1}^{N}\log\int p(\bm{x}_n|\bm{\theta})^{\alpha} q^{\setminus n}(\bm{\theta}) d\bm{\theta} ,
\label{eq:power_ep_energy}
\end{align}
where the constraint of the optimization problem changes to $(N-\alpha)\bm{\lambda}_q + \alpha \bm{\lambda}_0 =\sum_{n=1}^{N}\bm{\lambda}_{\setminus n}$.

The optimization problem in (\ref{eq:problem}) can be solved using a double-loop algorithm
\cite{heskes2002expectation,opper2005expectation}. This algorithm alternates
between an optimization of the cavity parameters $\{\bm\lambda_{\setminus n}\}$ in the
inner loop and an optimization of the parameters of the posterior approximation
$\bm{\lambda}_q$ in the outer loop. Each iteration of the double-loop algorithm is
guaranteed to minimize the energy in (\ref{eq:energy}). However, the
alternating optimization of $\bm{\lambda}_q$ and $\{\bm\lambda_{\setminus n}\}$ is very
inefficient to be useful in practice.

\section{Linear regression example}

In this section we demonstrate several properties of BB-$\alpha$ on a toy linear regression problem; in particular, we compare the BB-$\alpha$ optimal distribution to the true posterior in the cases where the true posterior lies in the variational family considered, and in the mean-field case where the variational family is Gaussian with diagonal covariance matrix.

More specifically, we consider a linear regression model of the form
\[
y_n = \bt^\top \bx_n + \sigma \varepsilon_n\,,
\]
where $y_n \in \mathbf{R}$, $\bx_n \in \mathbf{R}^2$ for $n=1,\ldots,N$, $\bt \in \mathbf{R}^2$, $\sigma \in \mathbf{R}_+$, and $(\varepsilon_n)_{n=1}^N \overset{\mathrm{iid}}{\sim} N(0,1)$. This model is simple enough to analytically optimize the BB-$\alpha$ energy function, and provides intuition for BB-$\alpha$ in more general contexts. We specify a prior $p_0(\bt) \propto \exp(-\bt^\top \bt/2)$ on the model weights; in this case, the posterior distribution for the model weights $\bt$ is given by
\begin{align*}
p(\bt) & \propto p_0(\bt) \prod_{n=1}^N p(y_n| \bt, \bx_n) \\
 & \propto \exp\left(-\frac{1}{2} \bt^\top \bt -\sum_{n=1}^N\frac{1}{2\sigma^2}\left( y_n - \bx_n^\top \bt \right)^2 \right) \\
 & \propto \exp\left(-\frac{1}{2} \bt^\top (I + \frac{1}{\sigma^2}\sum_{n=1}^N\bx_n \bx_n^\top )\bt + (\frac{1}{\sigma^2}\sum_{n=1}^N y_n \bx_n^\top )\bt    \right)
\end{align*}
and so the posterior is Gaussian with covariance matrix and mean given by
\begin{subequations}
	\label{eq:TrueMoments}
\begin{align}
\Sigma & = \left(I + \frac{1}{\sigma^2}\sum_{n=1}^N\bx_n \bx_n^\top\right)^{-1}\,, \\
\mu & = \left(I + \frac{1}{\sigma^2}\sum_{n=1}^N\bx_n \bx_n^\top\right)^{-1} \frac{1}{\sigma^2}\sum_{n=1}^N y_n \bx_n\,.
\end{align}
\end{subequations}

\subsection{Non-recovery of posterior distribution}

We first consider the case of a variational family of distributions which contains the true posterior distribution for this model. We let $f(\bt)$ be an arbitrary $2-$dimensional Gaussian distribution
\[
f(\bt) \propto \exp\left( -\frac{1}{2} \bt^\top \Lambda \bt + \eta^\top \bt \right)
\]
parameterized by its natural parameters $\Lambda \in \mathbf{R}^{2 \times 2}$ and $\eta \in \mathbf{R}^2$, and consider the variational family of the form $q(\bt) \propto p_0(\bt) f(\bt)^N$. As described in the main text, the BB-$\alpha$ optimality equations for the variational distribution $q$ are given by
\begin{align}\label{eq:MatchMoments}
\mathbf{E}_q\left\lbrack \mathbf{s}(\bt) \right\rbrack = \frac{1}{N} \sum_{n=1}^N \mathbf{E}_{\widetilde{p}_n}\left\lbrack \mathbf{s}(\bt) \right\rbrack\,,
\end{align}
where $\mathbf{s}$ is a vector of sufficient statistics for $q$, and $\widetilde{p}_n(\bt) \propto q(\bt) (p(y_n |\bt, \bx_n)/f(\bt))^\alpha$ is the tilted distribution for data point $n$.

Since the variational family considered is 2-dimensional Gaussian, the sufficient statistics for $q$ are $\mathbf{s}(\bt) = ((\theta_{i})_{i=1}^2, (\theta_i \theta_j)_{i, j = 1}^{2,2})$. We denote the solution of Equation \eqref{eq:MatchMoments} by $q^\star$, and denote its mean and variance by $\mu_{q^\star}$ and $\Sigma_{q^\star}$ respectively; we denote the corresponding quantities for the tilted distribution $\widetilde{p}_n$ by $\widetilde{\mu}_n$ and $\widetilde{\Sigma}_n$. Equation \eqref{eq:MatchMoments} then becomes
\begin{subequations}
	\begin{align}
	\mu_{q^\star} & = \frac{1}{N} \sum_{n=1}^N \widetilde{\mu}_n\,, \\
	\Sigma_{q^\star} + \mu_{q^\star}\mu_{q^\star}^\top & = \frac{1}{N} \left( \sum_{n=1}^N \widetilde{\Sigma}_n + \widetilde{\mu}_n \widetilde{\mu}_n^\top \right)\,.
	\end{align}
\end{subequations}
Or, in terms of natural parameters,
\begin{subequations} 	\label{eq:GMM}
	\begin{align}
	\Lambda_{q^\star}^{-1} \eta_{q^\star} & = \frac{1}{N}\sum_{n=1}^N \widetilde{\Lambda}^{-1}_{n} \widetilde{\eta}_{n}\,, \label{eq:GaussianMomentMatch} \\
	\Lambda_{q^\star}^{-1} + (\Lambda_{q^\star}^{-1} \eta_{q^\star})(\Lambda_{q^\star}^{-1} \eta_{q^\star})^\top & = \frac{1}{N} \sum_{n=1}^N \left(\widetilde{\Lambda}_{n}^{-1} + (\widetilde{\Lambda}_{n}^{-1} \widetilde{\eta}_{n})(\widetilde{\Lambda}_{n}^{-1} \eta_{n})^\top \right)\,. \label{eq:GaussianMomentMatch1}
	\end{align}
\end{subequations}

But the natural parameters of the true posterior are the mean of the natural parameters of the tilted distributions; i.e. power EP returns the true posterior when it lies in the approximating family. Since the map from moments to natural parameters is non-linear, we deduce that the distribution $q$ fitted according the moment matching equations \eqref{eq:GMM} is not equal to the true posterior for non-zero $\alpha$.

\subsection{Example 1}
We now provide a concrete example of this phenomenon, by performing the explicit calculations for a toy dataset consisting of two observations. We select our dataset to given by $\bx_1=(1,0)^\top$ and $\bx_2 = (0,1)^\top$, for now letting the output points $y_{1:2}$ be arbitrary. In this case, we can read off the mean and covariance of the true posterior from \eqref{eq:TrueMoments}:
\[
\Sigma = \frac{\sigma^2}{1+\sigma^2}I_2\, , \qquad \mu = \frac{1}{1+\sigma^2} (y_1, y_2)^\top\,.
\]
Note that in this case the true posterior has a diagonal covariance matrix. We consider fitting a variational family of distributions containing the true posterior via BB-$\alpha$, to demonstrate that the true posterior is generally not recovered for non-zero $\alpha$.

We now suppose that
\[
f(\bt) \propto \exp\left( -\frac{1}{2} \bt^\top \Lambda \bt + \eta^\top \bt \right)
\]
is constrained to have diagonal precision matrix $\Lambda$ (and hence diagonal covariance matrix). The optimality equation is still given by \eqref{eq:MatchMoments}, but the sufficient statistics are now given by the reduced vector $\mathbf{s}(\bt) = ((\theta_i)_{i=1}^2, (\theta_i^2)_{i=1}^2)$. Now, we have $q(\bt) \propto p_0(\bt)f(\bt)^N$, and so
\begin{align*}
q(\bt) & \propto \exp\left(-\frac{1}{2} \bt^\top \bt\right) \exp\left(-\frac{1}{2} \bt^\top (N\Lambda)\bt + N\eta^\top \bt\right) \\
 & = \exp\left(-\frac{1}{2} \bt^\top (I +N\Lambda ) \bt + N\eta^\top \bt\right)\,.
\end{align*}
Similarly, note that the tilted distribution $\widetilde{p}_n(\bt) \propto p_0(\bt) f(\bt)^{N-\alpha} p(y_n|\bt, \bx_n)^\alpha$. We note that
\begin{align*}
p(y_n|\bt, \bx_n) & \propto \exp\left(-\frac{1}{2\sigma^2}(y_n - \bx_n^\top \bt)^2\right)\\
 & \propto \exp\left( y_n\bx_n^\top \bt - \frac{1}{2\sigma^2} \bt^\top (\bx_n \bx_n^\top )\bt \right)\,.
\end{align*}
So we have
\begin{subequations}
\label{eq:TiltedDist}
\begin{align}
\widetilde{p}_n(\bt) & \propto \exp(-\frac{1}{2} \bt^\top \bt) \exp\left(-\frac{1}{2} \bt^\top ((N-\alpha)\Lambda)\bt + (N-\alpha)\eta^\top \bt\right) \exp\left( \frac{\alpha}{\sigma^2} y_n\bx_n^\top \bt - \frac{\alpha}{2 \sigma^2} \bt^\top (\bx_n \bx_n^\top )\bt \right) \\
&\propto \exp\left(-\frac{1}{2} \bt^\top \left(I + (N-\alpha)\Lambda + \frac{\alpha}{\sigma^2}\bx_n \bx_n^\top\right) \bt + \left( (N - \alpha) \eta^\top + \frac{\alpha}{\sigma^2} y_n \bx_n^\top \right)\bt \right)\,.
\end{align}
\end{subequations}
We can now use these expressions for the natural parameters, together with the optimality conditions \eqref{eq:MatchMoments} to fit the variational family of distributions.

\subsubsection{Matching first moments of the variational distribution}
Denoting $\Lambda = \mathrm{diag}(\lambda_1, \lambda_2)$, setting $N=2$ and using the specific values of $\bx_{1:2}$ mentioned above, the optimality equation \eqref{eq:GaussianMomentMatch} implies that we have
\[
\frac{2 \eta_i}{1 + 2\lambda_i} = \frac{1}{2}\left( \frac{(2-\alpha)\eta_i + \frac{\alpha}{\sigma^2}y_i}{1+(2-\alpha)\lambda_i + \frac{\alpha}{\sigma^2}}\right) + \frac{1}{2} \left( \frac{(2-\alpha) \eta_i}{1+(2-\alpha)\lambda_i} \right) \, , \qquad i=1,2\,.
\]
These equations are linear in the components of $\eta$, so we have that
\[
\eta_i = \frac{(y_i\frac{\alpha}{\sigma^2})/(1+(2-\alpha)\lambda_i + \frac{\alpha}{\sigma^2})}{ \frac{4}{1+2\lambda_i} - \frac{2-\alpha}{1+(2-\alpha)\lambda_i + \frac{\alpha}{\sigma^2}} - \frac{2-\alpha}{1+(2-\alpha)\lambda_i} }\, , \qquad i=1,2\,,
\]
yielding the natural parameter $\eta_i$ in terms of $\lambda_i$.

\subsubsection{Matching second moments of the variational distribution}
\begin{figure}[!htp]
	\centering
	\begin{minipage}[t]{0.45\textwidth}
		\includegraphics[width=\textwidth]{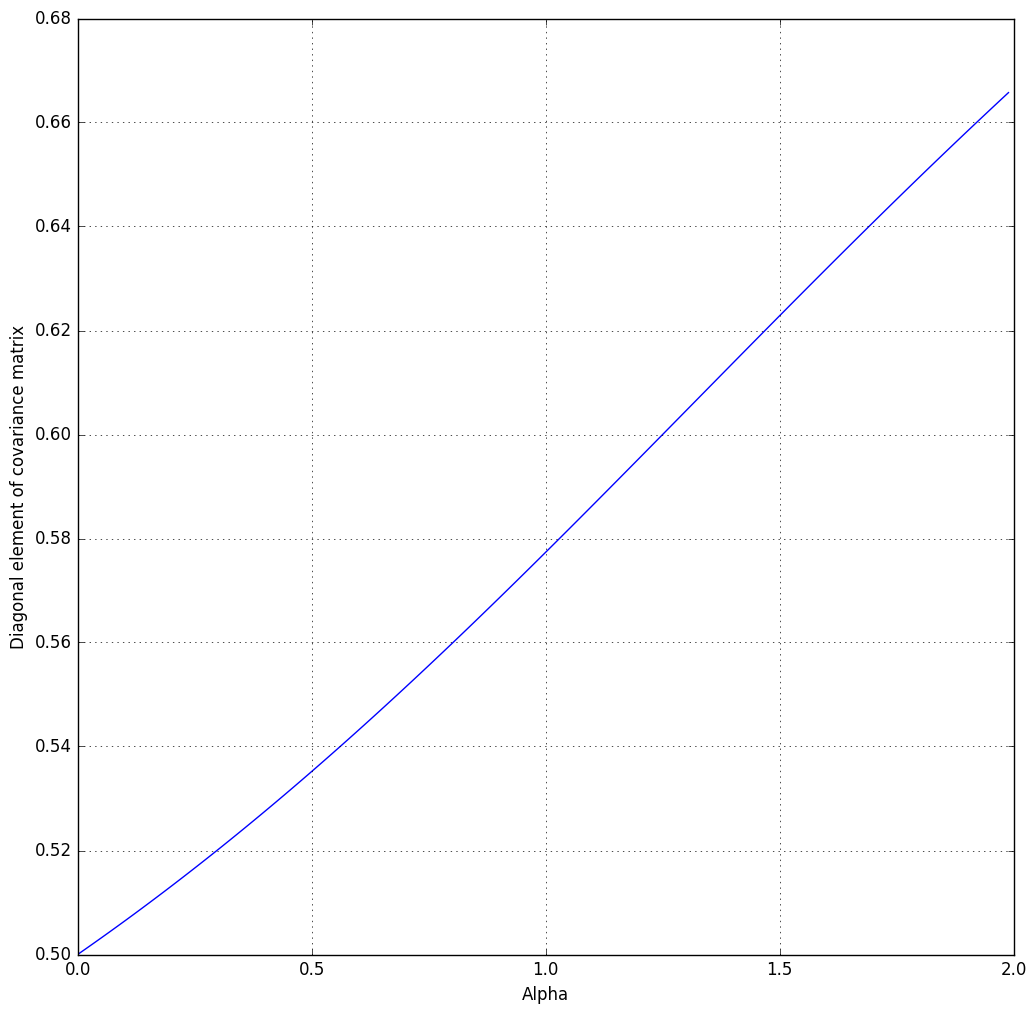}
		\caption{Diagonal element of fitted covariance matrix against $\alpha$, using output data points $y_1 = y_2 = 0$ and $\sigma^2=1$}
		\label{fig:VarVsAlpha1}
	\end{minipage}
	\hfill
	\begin{minipage}[t]{0.45\textwidth}
	\includegraphics[width=\textwidth]{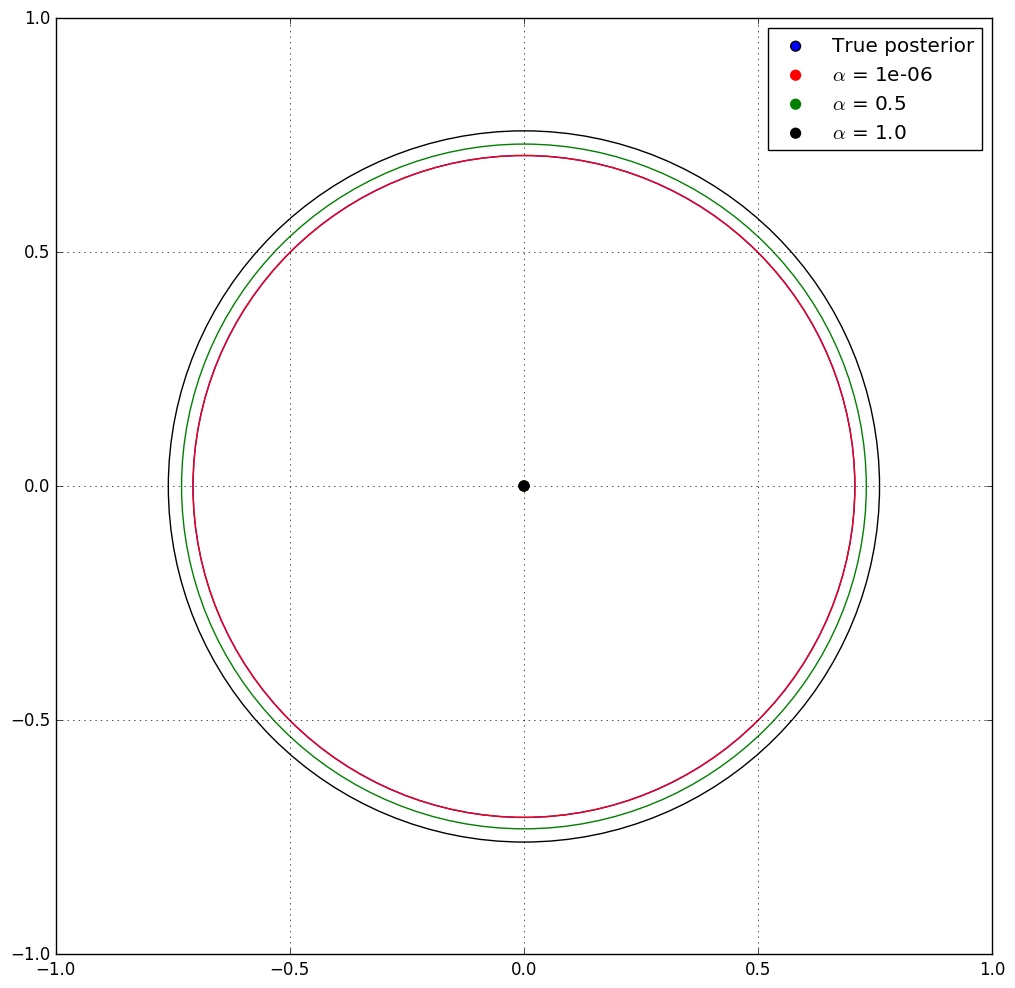}
	\caption{Plot of the mean and one standard deviation's confidence region for the true posterior and several BB-$\alpha$ approximations. Output data is set to $y_1 = 0$, $y_2 = 0$ and $\sigma^2 = 1$}
	\label{fig:true_vs_bba}
	\end{minipage}
\end{figure}

The optimality equation \eqref{eq:GaussianMomentMatch1} can now be used to recover the precision matrix parameters $\lambda_{1:2}$, and hence also the parameters $\eta_{1:2}$ using the formula derived in the above section. Importantly, recall that since we are dealing only with Gaussian variational distributions with diagonal covariance matrices, we need only the diagonal elements of this matrix equation. The resulting equations for $\lambda_{1:2}$ are quite involved, so we begin by evaluating the left-hand side of Equation \eqref{eq:GaussianMomentMatch1}. Evaluating the $i^\mathrm{th}$ diagonal element (where $i=1$ or $2$) gives
\[
\frac{1}{1 + 2\lambda_i} + \frac{4\eta_i^2}{(1 + 2\lambda_i)^2}\,.
\]
We now turn our attention to the right-hand side. We again evaluate the $i^{\mathrm{th}}$ diagonal element, which yields
\[
\frac{1}{2}\left( \frac{1}{1 + (2-\alpha)\lambda_i + \frac{\alpha}{\sigma^2}} + \left(\frac{(2-\alpha)\eta_i + \frac{\alpha}{\sigma^2}y_i}{1 + (2-\alpha)\lambda_i + \frac{\alpha}{\sigma^2}}\right)^2 \right) + \frac{1}{2}\left( \frac{1}{1 + (2-\alpha)\lambda_i} + \left(\frac{(2-\alpha)\eta_i}{1 + (2-\alpha)\lambda_i}\right)^2 \right)\,.
\]
Equating the above two expressions
%and introducing the notation
%\[
%p_1(\lambda) = 1 + 2\lambda\, , \ \ p_2(\lambda) = 1+(2-\alpha)\lambda \, , \ \ p_3(\lambda) = 1 + %(2-\alpha)\lambda + \frac{\alpha}{\sigma^2}\, , \ \ p_4(\lambda) = \alpha+2\sigma^2 + 2 + %2\sigma^2(2-\alpha)\lambda
%\]
in general yields $\lambda_i$ as a zero of a high-order polynomial, and therefore does not have an analytic solution. However, taking the data points $y_{1:2}$ to be zero simplifies the algebra considerably, and allows for an analytic solution for the natural parameters to be reached. In this case, the optimality equation yields
\[
\frac{1}{1+2\lambda_i} =  \frac{1}{2}\left(\frac{1}{1 + (2-\alpha)\lambda_i + \frac{\alpha}{\sigma^2}} + \frac{1}{1 + (2-\alpha)\lambda_i} \right) \, , \qquad i=1,2\,.
\]
Solving this fixed point equation (with the constraints that $1+2\lambda_i > 0$, $1 + (2-\alpha)\lambda_i + \frac{\alpha}{\sigma^2} > 0$ and $1 + (2-\alpha)\lambda_i > 0$ to make sure $q$ and $\tilde{p}_n$ are valid distributions) gives
\[
\lambda_i = \frac{\sqrt{\alpha^2 - 2\alpha + (\sigma^2 + 1)^2} - \alpha - \sigma^2 + 1}{2\sigma^2(2-\alpha)}  \, , \qquad i=1,2\,.
\]
Plotting a diagonal element of the variational covariance matrix as a function of $\alpha$ gives the curve plotted in Figure \ref{fig:VarVsAlpha1}. This plot demonstrates that the variance of the fitted distribution increases continuously and monotonically with $\alpha$ in the range $(0,2)$.

With the fitted mean and covariance matrix, we can plot the fitted approximate posterior distributions as $\alpha$ varies, along with the true posterior, see Figure \ref{fig:true_vs_bba}. Note that in the limit as $\alpha \rightarrow 0$, the true posterior is recovered by BB-$\alpha$, as the corresponding $\alpha$-divergence converges to the KL-divergence; in fact, the true posterior and the fitted approximation for $\alpha = 10^{-6}$ are indistinguishable at this scale. Note that for non-zero $\alpha$, the fitted covariance parameters differ from those of the true posterior.

\subsection{Mean-field approximations}
We now demonstrate the behavior of BB-$\alpha$ in fitting mean-field Gaussian distributions, showing that smooth interpolation between KL-like and EP-like behavior can be obtained by varying $\alpha$.

\subsection{Example 2}
We now consider the case where the true posterior has non-diagonal covariance matrix; we set the input values in our toy dataset to be $\bx_1 = (1,-1)^\top$, $\bx_2 = (-1,1)^\top$ (again leaving the output data $y_{1:2}$ arbitrary for now), and note from Equation \eqref{eq:TrueMoments} that this implies the mean and covariance of the true posterior are given by
\begin{align*}
\Sigma & = \frac{1}{4 + \sigma^2}\begin{pmatrix} \sigma^2 + 2 & 2 \\ 2 & \sigma^2 + 2 \end{pmatrix}\,, \\
 \mu & = \left( \frac{y_1 - y_2}{4 + \sigma^2}, \frac{y_2 - y_1}{4 + \sigma^2} \right)^\top\,.
\end{align*}
We fit the same variational family considered in Example 1; namely the family in which
\[
f(\bt) \propto \exp\left( -\frac{1}{2} \bt^\top \Lambda \bt + \eta^\top \bt \right)
\]
is constrained to have diagonal precision matrix $\Lambda$ (and hence covariance matrix). We can now use this information, together with the form of the tilted distributions in Equation \eqref{eq:TiltedDist} and the optimality conditions in Equation \eqref{eq:MatchMoments} to fit the variational family of distributions.

\subsubsection{Matching first moments of the variational distribution}
Denoting $\Lambda = \mathrm{diag}(\lambda_1, \lambda_2)$, substituting in $N=2$ and using our specific choice of data points $\bx_{1:2}$, Equation \eqref{eq:GaussianMomentMatch} yields the linear system
\[
\begin{pmatrix}\frac{2}{1+2\lambda_1} & 0 \\ 0 & \frac{2}{1+2\lambda_2} \end{pmatrix} \eta = \begin{pmatrix}1+(2-\alpha)\lambda_1 + \frac{\alpha}{\sigma^2} & -\frac{\alpha}{\sigma^2} \\ -\frac{\alpha}{\sigma^2} & 1+(2-\alpha)\lambda_2 + \frac{\alpha}{\sigma^2} \end{pmatrix}^{-1}\left( (2-\alpha) \eta +\frac{\alpha}{2\sigma^2}\begin{pmatrix} y_1 -y_2 \\ y_2 - y_1 \end{pmatrix} \right)\,.
\]
This linear system can be solved for $\eta$, although in general the solution is a complicated rational function of $(\lambda_1, \lambda_2)$. However, taking $y_1 = y_2$ yields $\eta = 0$.

\subsubsection{Matching the second moments of the variational distribution}
\begin{figure}[!htp]
	\centering
	\begin{minipage}[t]{0.45\textwidth}
		\includegraphics[width=\textwidth]{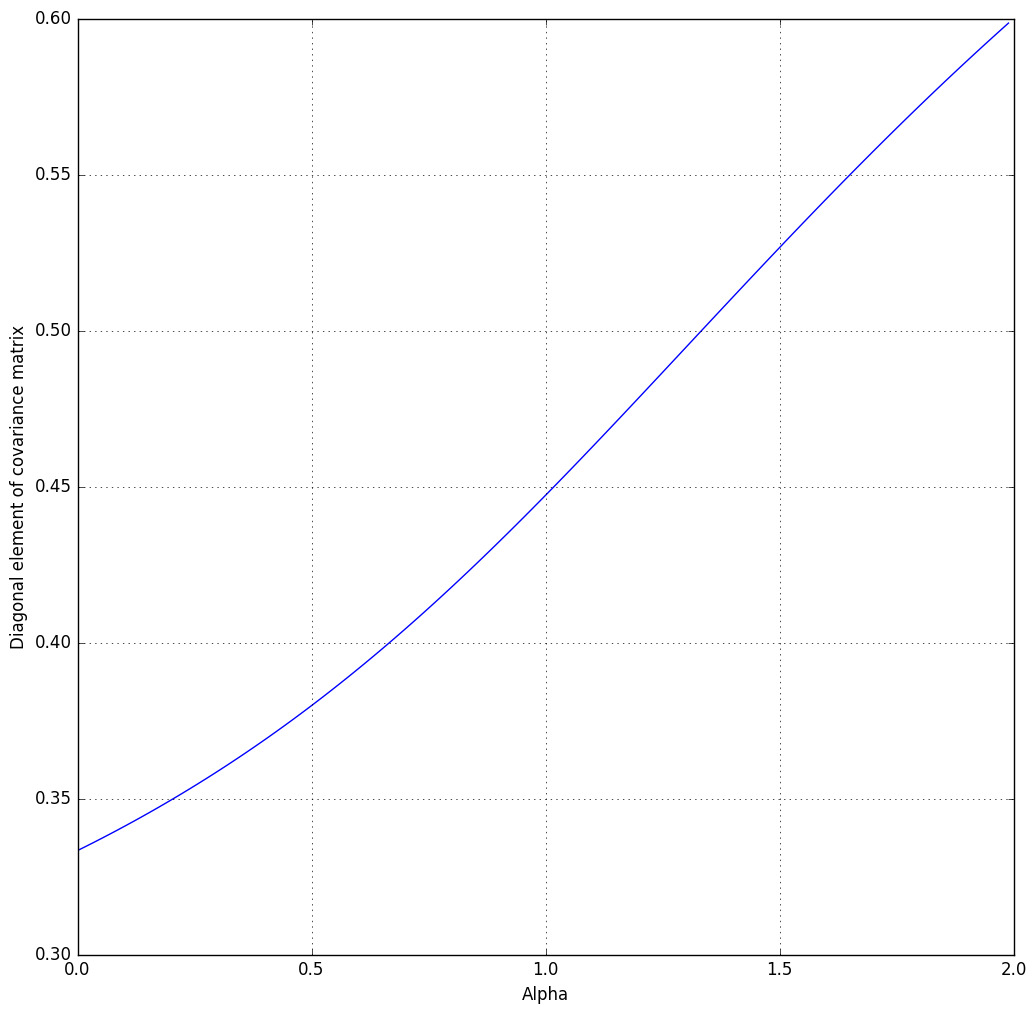}
		\caption{Diagonal element of fitted covariance matrix against $\alpha$, using output data points $y_1 = 0$,  $y_2 = 0$ and $\sigma^2=1$}
		\label{fig:VarVsAlphaAlt1}
	\end{minipage}
	\hfill
	\begin{minipage}[t]{0.45\textwidth}
		\includegraphics[width=\textwidth]{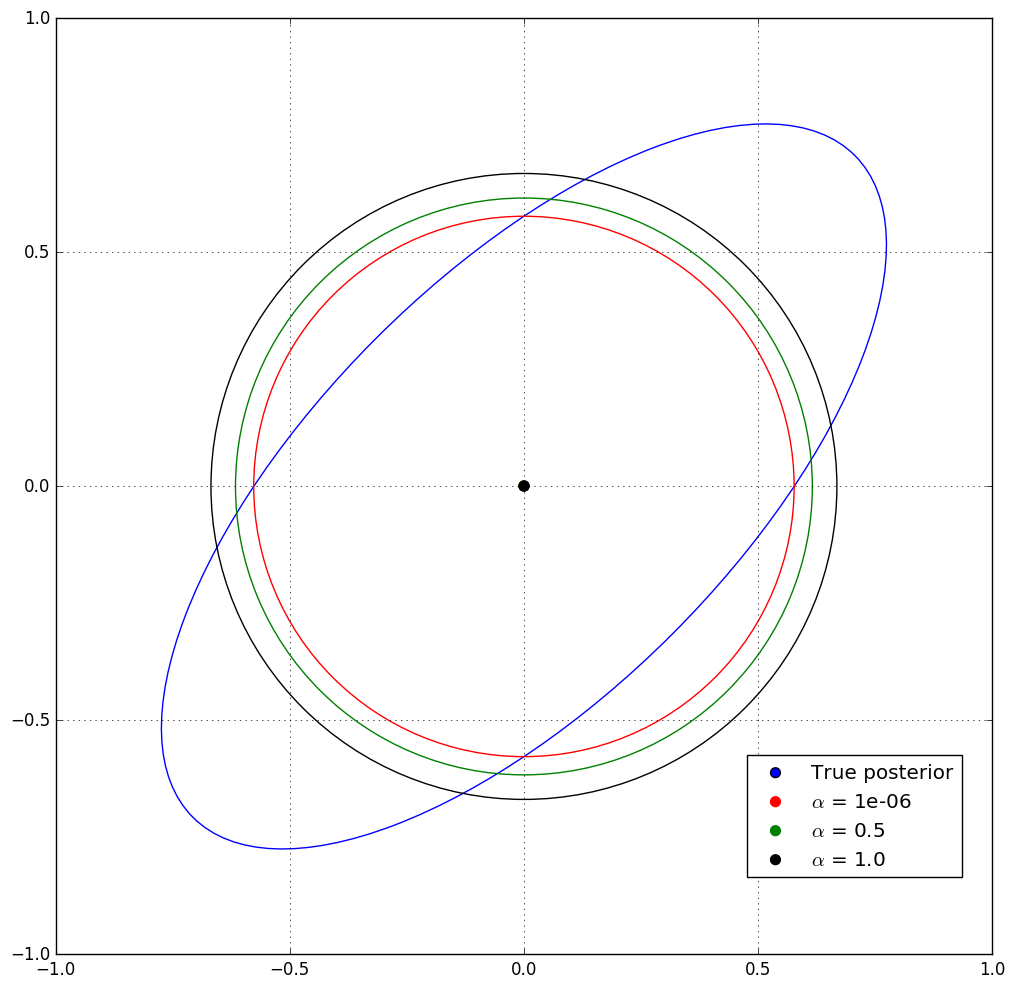}
		\caption{Plot of mean and one standard deviation's confidence region for the true posterior and several BB-$\alpha$ approximations. Output data is set to $y_1 = 0$, $y_2 = 0$ and $\sigma^2 = 1$}
		\label{fig:true_vs_bba2}
	\end{minipage}
\end{figure}

With the above choices for $y_1$ and $y_2$, it follows by symmetry that we must have $\lambda_1 = \lambda_2$. We denote this unknown variable by $\lambda$ in what follows. Considering the diagonal elements of the Equation \eqref{eq:GaussianMomentMatch1} then yields
\[
\frac{1}{1+2\lambda_i} = \frac{1 + (2-\alpha)\lambda_i + \alpha/\sigma^2}{(1+(2-\alpha)\lambda_i + \alpha/\sigma^2) - \alpha^2/\sigma^2} + \frac{y^2 \alpha^2}{\sigma^4}\frac{(1 + (2-\alpha)\lambda_i)^2}{(1 + (2-\alpha)\lambda_i + \alpha/\sigma^2)^2 - \alpha^2/\sigma^4}\,.
\]
This can be re-arranged into a cubic for $\lambda$ and thus solved analytically, at least in theory; the resulting expression for $\lambda$ in terms of $\alpha, \sigma$ and $y$ is lengthy in practice and is therefore omitted here. We instead consider the case $y = y_1 = y_2 = 0$, where the algebra is more tractable. This results in the following equation for $\lambda$:
\[
\frac{1}{1 + 2\lambda_i} = \frac{1 + (2-\alpha)\lambda_i + \alpha/\sigma^2}{(1+(2-\alpha)\lambda_i + \alpha/\sigma^2)^2 - \alpha^2/\sigma^4} \, , \qquad i=1,2\,.
\]
This equation is merely quadratic, and hence has a more easily expressible solution; solving this equation (with the constraint that both sides are positive, and the denominator of the RHS is also positive) gives the precision parameters as
\[
\lambda_i = \frac{\sqrt{4\alpha^2 -8\alpha + \sigma^4 +4\sigma^2 + 4} - (2\alpha + \sigma^2 - 2)}{2\sigma^2(2 - \alpha)}\, , \qquad i = 1,2\,.
\]
Plotting a diagonal element of the corresponding fitted covariance matrix as a function of $\alpha$ gives the curve shown in Figure \ref{fig:VarVsAlphaAlt1}. Note that as before, the element exhibits a continuous, monotonic dependence on $\alpha$ in the range $(0,2)$.

With the fitted mean and covariance matrix, we can plot the fitted approximate posterior distributions as $\alpha$ varies, along with the true posterior, see Figure \ref{fig:true_vs_bba2}. Note that in the limit as $\alpha \rightarrow 0$, the BB-$\alpha$ solution mimics the KL fit, exhibiting low variance relative to the true posterior, and as $\alpha$ increases, so does the spread of the distribution, consistent with Figure \ref{fig:VarVsAlphaAlt1}.

\section{Results on toy dataset with neural network regression}

\begin{figure}
\includegraphics[width=0.24\textwidth]{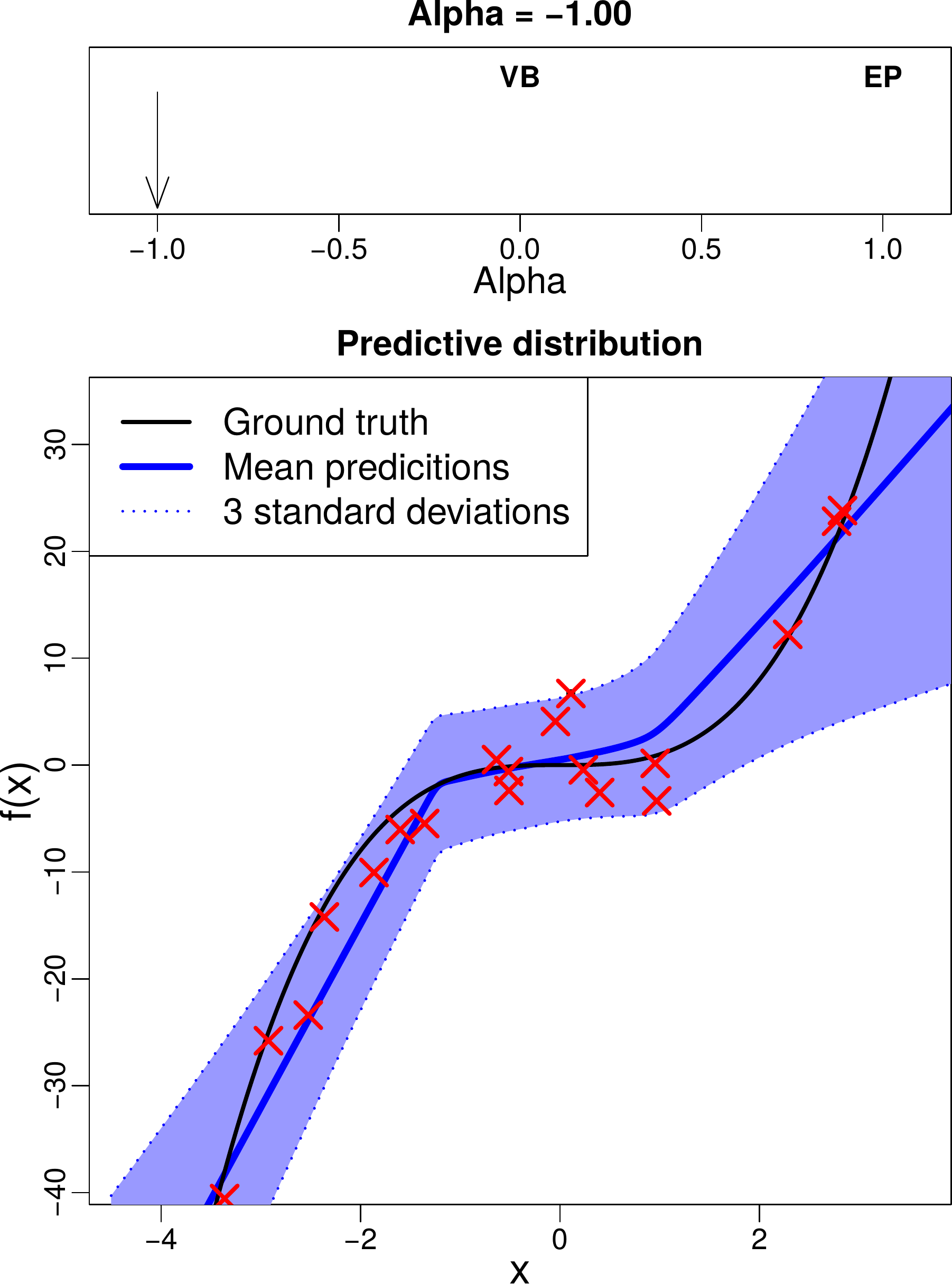}
\includegraphics[width=0.24\textwidth]{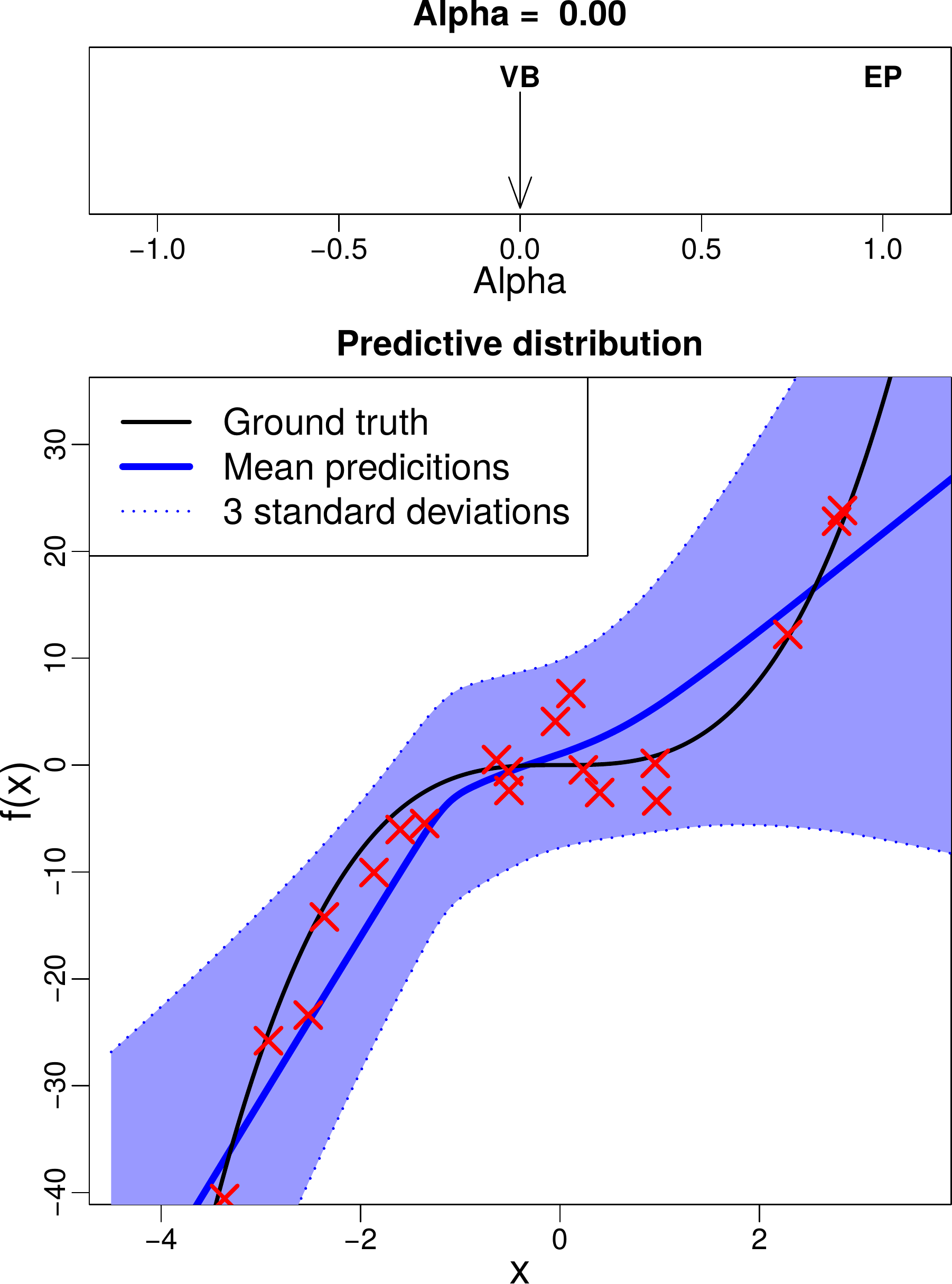}
\includegraphics[width=0.24\textwidth]{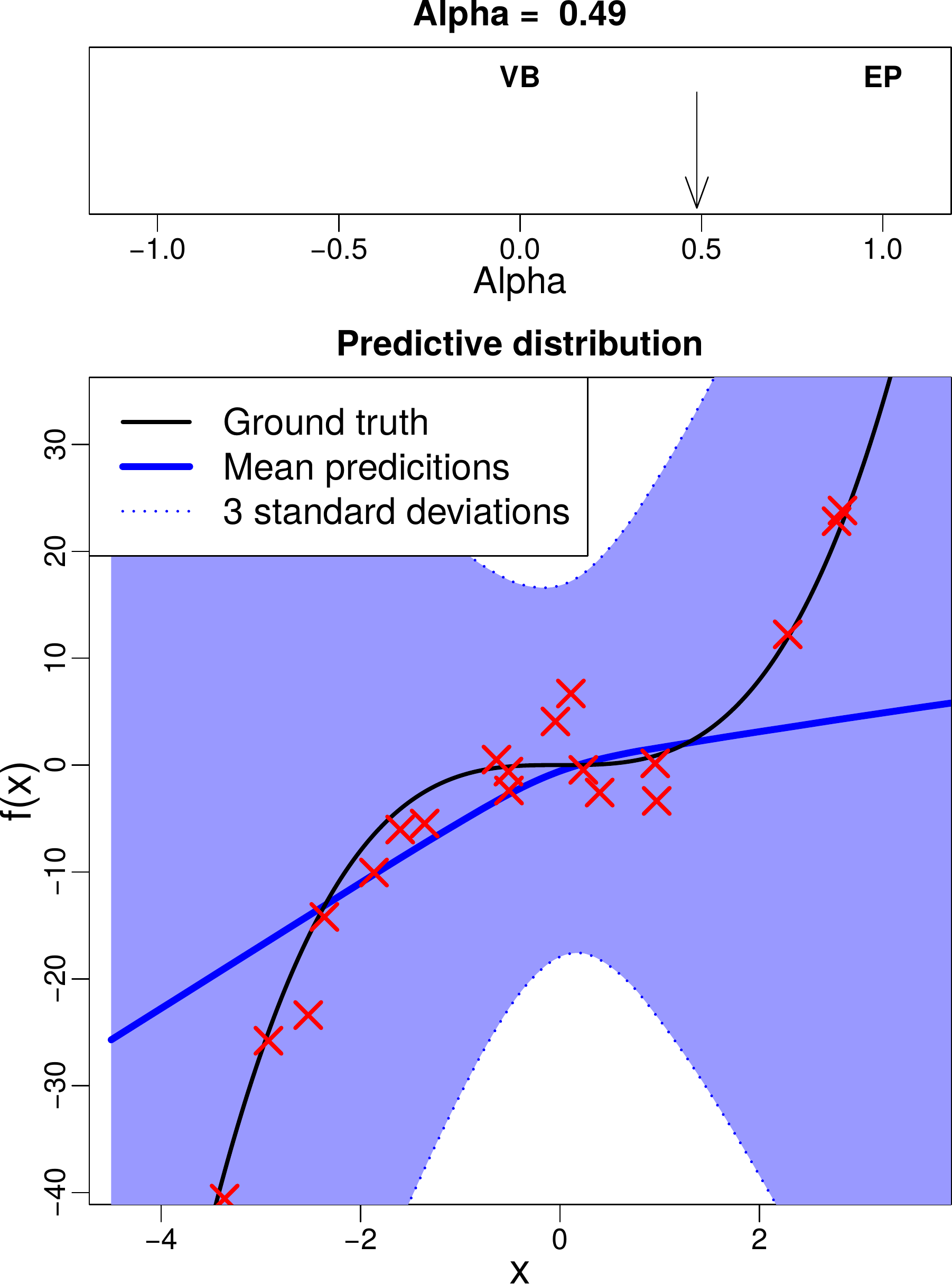}
\includegraphics[width=0.24\textwidth]{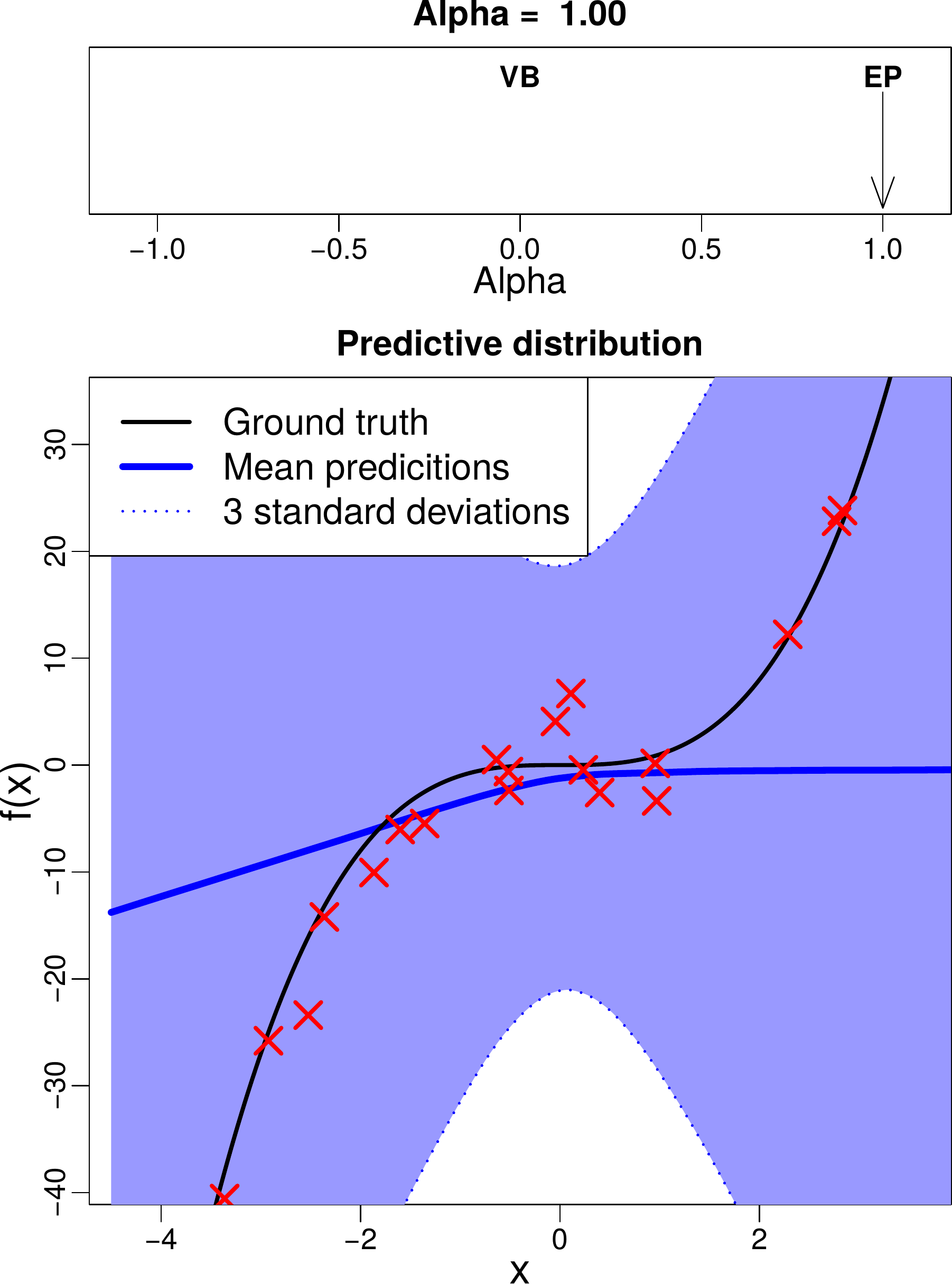}
\caption{Predictions obtained for different values of $\alpha$ in the toy dataset with neural networks.}\label{fig:toy_nnet_example}
\end{figure}

We evaluated the predictions obtained by neural networks
trained with BB-$\alpha$ in the toy dataset described in
\cite{Hernandez-Lobato15b}.  This dataset is generated by sampling 20
inputs~$x$ uniformly at random in the interval~$[-4,4]$.  For each value of~$x$
obtained, the corresponding target $y$ is computed as~${y = x^3 + \epsilon_n}$,
where~${\epsilon_n \sim \mathcal{N}(0, 9)}$. We fitted a neural network with
one hidden layer with 100 hidden units and rectifier activation functions. In
this neural network model, we fixed the variance of the output noise optimally
to $\sigma^2 = 9$ and kept the prior variance for the weights fixed to value
one. Figure \ref{fig:toy_nnet_example} shows plots of the predictive
distributions obtained for different values of $\alpha$. We can see that
smaller and negative values of $\alpha$ produce lower variance in the
predictive distributions, while larger values of $\alpha$ result in higher
predictive variance.

%\bibliographystyle{apalike}
%\bibliography{../references.bib}
%
%
%\end{document}

%% file: bb_alpha_icml.bbl
\begin{thebibliography}{31}
\providecommand{\natexlab}[1]{#1}
\providecommand{\url}[1]{\texttt{#1}}
\expandafter\ifx\csname urlstyle\endcsname\relax
  \providecommand{\doi}[1]{doi: #1}\else
  \providecommand{\doi}{doi: \begingroup \urlstyle{rm}\Url}\fi

\bibitem[Amari(1985)]{amari1985}
Amari, Shun-ichi.
\newblock \emph{Differential-Geometrical Methods in Statistic}.
\newblock Springer, New York, 1985.

\bibitem[Barthelm{\'e} \& Chopin(2011)Barthelm{\'e} and
  Chopin]{barthelme2011abc}
Barthelm{\'e}, Simon and Chopin, Nicolas.
\newblock Abc-ep: Expectation propagation for likelihoodfree {B}ayesian
  computation.
\newblock In \emph{ICML}, 2011.

\bibitem[Bastien et~al.(2012)Bastien, Lamblin, Pascanu, Bergstra, Goodfellow,
  Bergeron, Bouchard, and Bengio]{Bastien-Theano-2012}
Bastien, Fr{\'{e}}d{\'{e}}ric, Lamblin, Pascal, Pascanu, Razvan, Bergstra,
  James, Goodfellow, Ian~J., Bergeron, Arnaud, Bouchard, Nicolas, and Bengio,
  Yoshua.
\newblock Theano: new features and speed improvements.
\newblock Deep Learning and Unsupervised Feature Learning NIPS 2012 Workshop,
  2012.

\bibitem[Bottou(1998)]{bottou1998}
Bottou, L{\'e}on.
\newblock Online learning and stochastic approximations.
\newblock \emph{On-line learning in neural networks}, 17\penalty0 (9):\penalty0
  25, 1998.

\bibitem[Cunningham et~al.(2011)Cunningham, Hennig, and
  Lacoste-Julien]{cunningham2011}
Cunningham, John~P, Hennig, Philipp, and Lacoste-Julien, Simon.
\newblock Gaussian probabilities and expectation propagation.
\newblock \emph{arXiv preprint arXiv:1111.6832}, 2011.

\bibitem[Dehaene \& Barthelm\'e(2015)Dehaene and Barthelm\'e]{dehaene2015}
Dehaene, Guillaume and Barthelm\'e, Simon.
\newblock Expectation propagation in the large-data limit.
\newblock \emph{arXiv:1503.08060}, 2015.

\bibitem[Gelman et~al.(2014)Gelman, Vehtari, Jylänki, Robert, Chopin, and
  Cunningham]{gelman2014}
Gelman, Andrew, Vehtari, Aki, Jylänki, Pasi, Robert, Christian, Chopin,
  Nicolas, and Cunningham, John~P.
\newblock Expectation propagation as a way of life.
\newblock \emph{arXiv:1412.4869}, 2014.

\bibitem[Glorot \& Bengio(2010)Glorot and Bengio]{glorot2010understanding}
Glorot, Xavier and Bengio, Yoshua.
\newblock Understanding the difficulty of training deep feedforward neural
  networks.
\newblock In \emph{AISTATS}, pp.\  249--256, 2010.

\bibitem[Hachmann et~al.(2014)Hachmann, Olivares-Amaya, Jinich, Appleton,
  Blood-Forsythe, Seress, Rom{\'a}n-Salgado, Trepte, Atahan-Evrenk, Er,
  et~al.]{hachmann2014lead}
Hachmann, Johannes, Olivares-Amaya, Roberto, Jinich, Adrian, Appleton,
  Anthony~L, Blood-Forsythe, Martin~A, Seress, L{\'a}szl{\'o}~R,
  Rom{\'a}n-Salgado, Carolina, Trepte, Kai, Atahan-Evrenk, Sule, Er,
  S{\"u}leyman, et~al.
\newblock Lead candidates for high-performance organic photovoltaics from
  high-throughput quantum chemistry--the harvard clean energy project.
\newblock \emph{Energy \& Environmental Science}, 7\penalty0 (2):\penalty0
  698--704, 2014.

\bibitem[Hern{\'a}ndez-Lobato \& Hern{\'a}ndez-Lobato(2016)Hern{\'a}ndez-Lobato
  and Hern{\'a}ndez-Lobato]{hernandez2016scalable}
Hern{\'a}ndez-Lobato, Daniel and Hern{\'a}ndez-Lobato, Jos{\'e}~Miguel.
\newblock Scalable gaussian process classification via expectation propagation.
\newblock In \emph{AISTATS}, 2016.

\bibitem[Heskes \& Zoeter(2002)Heskes and Zoeter]{heskes2002expectation}
Heskes, Tom and Zoeter, Onno.
\newblock Expectation propagation for approximate inference in dynamic
  {B}ayesian networks.
\newblock In \emph{UAI}, pp.\  216--223. Morgan Kaufmann Publishers Inc., 2002.

\bibitem[Jordan et~al.(1999)Jordan, Ghahramani, Jaakkola, and Saul]{jakkola99}
Jordan, M.~I., Ghahramani, Z., Jaakkola, T.~S., and Saul, L.~K.
\newblock An introduction to variational methods for graphical models.
\newblock \emph{Machine Learning}, 37:\penalty0 183--233, 1999.

\bibitem[Kingma \& Welling(2014)Kingma and Welling]{Kingma2014}
Kingma, D.~P. and Welling, M.
\newblock Auto-encoding variational bayes.
\newblock In \emph{ICLR}, 2014.

\bibitem[Kingma \& Ba(2014)Kingma and Ba]{kingma2014adam}
Kingma, Diederik and Ba, Jimmy.
\newblock Adam: A method for stochastic optimization.
\newblock \emph{arXiv preprint arXiv:1412.6980}, 2014.

\bibitem[Li et~al.(2015)Li, Hernandez-Lobato, and Turner]{li2015stochastic}
Li, Yingzhen, Hernandez-Lobato, Jose~Miguel, and Turner, Richard~E.
\newblock Stochastic expectation propagation.
\newblock In \emph{NIPS}, pp.\  2323--2331, 2015.

\bibitem[Lichman(2013)]{Lichman:2013}
Lichman, M.
\newblock {UCI} machine learning repository, 2013.
\newblock URL \url{http://archive.ics.uci.edu/ml}.

\bibitem[Minka(2001{\natexlab{a}})]{minka2001}
Minka, Thomas~P.
\newblock Expectation propagation for approximate {B}ayesian inference.
\newblock In \emph{UAI}, pp.\  362--369. Morgan Kaufmann Publishers Inc.,
  2001{\natexlab{a}}.

\bibitem[Minka(2001{\natexlab{b}})]{minka2001thesis}
Minka, Thomas~P.
\newblock \emph{A Family of Algorithms for Approximate {B}ayesian Inference}.
\newblock PhD thesis, Cambridge, MA, USA, 2001{\natexlab{b}}.
\newblock AAI0803033.

\bibitem[Minka(2004)]{minka2004}
Minka, Thomas~P.
\newblock Power ep.
\newblock Technical report, Technical report, Microsoft Research, Cambridge,
  2004.

\bibitem[Minka(2005)]{minka2005}
Minka, Thomas~P.
\newblock Divergence measures and message passing.
\newblock Technical report, Technical report, Microsoft Research, 2005.

\bibitem[Opper \& Winther(2005)Opper and Winther]{opper2005expectation}
Opper, Manfred and Winther, Ole.
\newblock Expectation consistent approximate inference.
\newblock \emph{The Journal of Machine Learning Research}, 6:\penalty0
  2177--2204, 2005.

\bibitem[Pyzer-Knapp et~al.(2015)Pyzer-Knapp, Li, and
  Aspuru-Guzik]{pyzer2015learning}
Pyzer-Knapp, Edward~O, Li, Kewei, and Aspuru-Guzik, Alan.
\newblock Learning from the harvard clean energy project: The use of neural
  networks to accelerate materials discovery.
\newblock \emph{Advanced Functional Materials}, 25\penalty0 (41):\penalty0
  6495--6502, 2015.

\bibitem[Ranganath et~al.(2014)Ranganath, Gerrish, and
  Blei]{ranganath2014black}
Ranganath, Rajesh, Gerrish, Sean, and Blei, David.
\newblock Black box variational inference.
\newblock In \emph{AISTATS}, pp.\  814--822, 2014.

\bibitem[Salimans et~al.(2013)Salimans, Knowles, et~al.]{salimans2013fixed}
Salimans, Tim, Knowles, David~A, et~al.
\newblock Fixed-form variational posterior approximation through stochastic
  linear regression.
\newblock \emph{{B}ayesian Analysis}, 8\penalty0 (4):\penalty0 837--882, 2013.

\bibitem[Seeger(2005)]{seeger2005}
Seeger, Matthias.
\newblock Expectation propagation for exponential families.
\newblock Technical report, 2005.

\bibitem[Teh et~al.(2015)Teh, Hasenclever, Lienart, Vollmer, Webb,
  Lakshminarayanan, and Blundell]{teh2015}
Teh, Yee~Whye, Hasenclever, Leonard, Lienart, Thibaut, Vollmer, Sebastian,
  Webb, Stefan, Lakshminarayanan, Balaji, and Blundell, Charles.
\newblock Distributed {B}ayesian learning with stochastic natural-gradient
  expectation propagation and the posterior server.
\newblock \emph{arXiv:1512.09327}, 2015.

\bibitem[Turner \& Sahani(2011{\natexlab{a}})Turner and Sahani]{Turner2011}
Turner, Richard~E. and Sahani, Maneesh.
\newblock Two problems with variational expectation maximisation for time
  series models.
\newblock In \emph{{B}ayesian Time Series Models}, pp.\  104--124. Cambridge
  University Press, 2011{\natexlab{a}}.
\newblock Cambridge Books Online.

\bibitem[Turner \& Sahani(2011{\natexlab{b}})Turner and
  Sahani]{turner+sahani2011}
Turner, Richard~E. and Sahani, Maneesh.
\newblock Probabilistic amplitude and frequency demodulation.
\newblock In \emph{NIPS}, pp.\  981--989. 2011{\natexlab{b}}.

\bibitem[Winn \& Bishop(2005)Winn and Bishop]{winn2005}
Winn, John~M and Bishop, Christopher~M.
\newblock Variational message passing.
\newblock In \emph{Journal of Machine Learning Research}, pp.\  661--694, 2005.

\bibitem[Xu et~al.(2014)Xu, Lakshminarayanan, Teh, Zhu, and Zhang]{xu2014}
Xu, Minjie, Lakshminarayanan, Balaji, Teh, Yee~Whye, Zhu, Jun, and Zhang, Bo.
\newblock Distributed {B}ayesian posterior sampling via moment sharing.
\newblock In \emph{NIPS}, pp.\  3356--3364, 2014.

\bibitem[Zhu \& Rohwer(1995)Zhu and Rohwer]{zhu1995}
Zhu, Huaiyu and Rohwer, Richard.
\newblock Information geometric measurements of generalisation.
\newblock Technical report, Technical Report NCRG/4350. Aston University.,
  1995.

\end{thebibliography}
